\documentclass[twocolumn]{article}

\usepackage{arxiv}

\usepackage[utf8]{inputenc} 
\usepackage[T1]{fontenc}    
\usepackage{hyperref}       
\usepackage{url}            
\usepackage{booktabs}       
\usepackage{amsfonts}       
\usepackage{nicefrac}       
\usepackage{microtype}      
\usepackage{lipsum}		
\usepackage{graphicx}
\usepackage{amsmath,amssymb}
\usepackage{booktabs}
\usepackage{longtable}
\usepackage{tabularx}
\usepackage{multirow}
\usepackage{xcolor}
\usepackage{algorithm}
\usepackage[numbers]{natbib}
\usepackage{doi}
\usepackage{algpseudocode}
\usepackage{comment}
\usepackage{tikz}
\usepackage{tabularx}
\usetikzlibrary{arrows.meta,positioning,fit,calc,shapes.geometric}

\algrenewcommand\algorithmicrequire{\textbf{Input:}}
\algrenewcommand\algorithmicensure{\textbf{Output:}}
\makeatletter
\newenvironment{breakablealgorithm}
  {%
   \begin{center}%
     \refstepcounter{algorithm}%
     \hrule height.8pt depth0pt \kern2pt%
     \renewcommand{\caption}[2][\relax]{%
       {\raggedright\textbf{\ALG@name~\thealgorithm} ##2\par}%
       \ifx\relax##1\relax
         \addcontentsline{loa}{algorithm}{\protect\numberline{\thealgorithm}##2}%
       \else
         \addcontentsline{loa}{algorithm}{\protect\numberline{\thealgorithm}##1}%
       \fi
       \kern2pt\hrule\kern2pt
     }%
  }%
  {%
     \kern2pt\hrule\relax%
   \end{center}%
  }%
\makeatother

\title{Spectral Model eXplainer: a chemically-grounded explainability framework for spectral-based machine learning models}

\author{ 
	\href{https://orcid.org/0000-0002-2744-4467}{\includegraphics[scale=0.06]{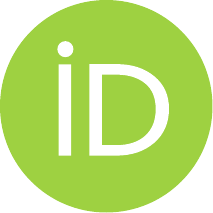}\hspace{1mm}José Vinícius Ribeiro}\thanks{Corresponding author} \\
	\small Applied Nuclear Physics Laboratory\\ \small State University of Londrina\\
	\small Dept. of Engineering and Architecture\\ \small University of Trieste\\
	\small \texttt{ribeirojosevinicius@gmail.com} \\
	\And
	\href{https://orcid.org/0000-0003-0651-1103}{\includegraphics[scale=0.06]{orcid.pdf}\hspace{1mm}Rafael Figueira Goncalves} \\
	\small Dept. of Mathematics and Geosciences\\ \small University of Trieste\\
	\small \texttt{rafael.figueiragoncalves@phd.units.it} \\
	\AND
	\href{https://orcid.org/0000-0003-1154-0339}{\includegraphics[scale=0.06]{orcid.pdf}\hspace{1mm}Fábio Luiz Melquiades} \\
	\small Applied Nuclear Physics Laboratory\\ \small State University of Londrina\\
	\small \texttt{fmelquiades@uel.br} \\
	\And
	\href{https://orcid.org/0000-0002-4988-0702}{\includegraphics[scale=0.06]{orcid.pdf}\hspace{1mm}Sylvio Barbon Junior} \\
	\small Dept. of Engineering and Architecture\\ \small University of Trieste\\
	\small \texttt{sylvio.barbonjunior@units.it} \\
}

\renewcommand{\shorttitle}{Spectral Model eXplainer}

\hypersetup{
pdftitle={Spectral Model eXplainer: a chemically-grounded explainability framework for spectral-based machine learning models},
pdfsubject={Computer Science, Machine Learning},
pdfauthor={José Vinícius Ribeiro, Rafael Figueira Goncalves, Fábio Luiz Melquiades, Sylvio Barbon Junior},
pdfkeywords={Machine learning, Chemometrics, eXplainable Artificial Intelligence (XAI), Explainability, Spectroscopy, Spectral zones, graph-based explainability},
}

\begin{document}

\twocolumn[
\maketitle
\begin{abstract}
Spectral-based machine learning models have been increasingly deployed in chemometrics and spectroscopy, where predictive accuracy is as important as explainability. Current employed eXplainable Artificial Intelligence (XAI) methods are largely adapted from tabular or generic multivariate domains, assigning relevance to isolated spectral variables rather than to the chemically meaningful spectral zones. Widely adopted tools such as SHapley Additive exPlanations (SHAP), Permutation Feature Importance (PFI), and Variable Importance in Projection scores (VIP) were not designed for the physical continuity and high collinearity of spectral data, and their variable-level outputs require post-hoc aggregation to recover zone-level information. This study introduces the Spectral Model eXplainer (SMX), a post-hoc, global, model-agnostic XAI framework that explains spectral classifiers through expert-informed spectral zones. SMX summarizes each zone via PCA, defines quantile-based logical predicates, estimates predicate relevance with perturbation in stochastic subsamples, and aggregates bag-wise rankings in a directed weighted graph summarized by Local Reaching Centrality. A key component is threshold spectrum reconstruction, which back-projects predicate boundaries to the original spectral domain in natural measurement units, enabling direct visual comparison with measured spectra. The method was evaluated on eight real spectral datasets (six based on X-ray Fluorescence--XRF and two based on Gamma-ray Spectrometry) and one synthetic benchmark with known ground truth, explaining classifiers with increasing complexity, namely PLS, SVM, and MLP. Comparative analyses via Wilcoxon signed-rank tests covering XAI desiderata indicated broad SMX faithfulness equivalence with baselines (SHAP, PFI, and VIP), competitive-to-superior domain alignment and stability, and simpler outputs than PFI and VIP while remaining comparable to SHAP. A soil fertility XRF-centered study case further showed that predicate-derived thresholds map model behavior to actionable spectral boundaries linked to meaningful elemental signals. Overall, by bridging the gap between model behavior and spectral interpretation, SMX represents a promising step toward spectral-native explainability, opening new avenues for integrating XAI insights into practical workflows, decision support, and physicochemical reasoning across spectroscopic applications. 
\end{abstract}

\vspace{0.5em}

\vspace{2em}
]

\clearpage

\section{Introduction}

Spectroscopy stands as one of the most versatile analytical tools in modern science. Grounded in the interaction between electromagnetic radiation and matter, it allows for the non-destructive, rapid, and often minimal or preparation-free characterization of the chemical and structural patterns of diverse sample matrices. Methods such as X-Ray Fluorescence (XRF), Gamma-Ray Spectrometry (GRS), visible and Near-InfraRed (vis-NIR), and Laser-Induced Breakdown Spectroscopy (LIBS) have found widespread applications across disciplines such as agriculture, food science, pharmaceuticals, materials science, medicine, among others. This ubiquity stems from a foundational principle: the spectral profile of a sample carries its characteristic, unique signature, encoding meaningful information that describes its identity and properties.

Despite such wealth of features contained in spectra, their complex, high-dimensional nature poses significant challenges for traditional, univariate statistical methods. In this context, coupling spectroscopic data with supervised Machine Learning (ML) has become a routine approach for high-throughput classification and regression tasks in analytical chemistry and chemometrics. As spectroscopic modeling increasingly influences analytical decisions, deploying such models in regulated or scientifically demanding contexts requires more than predictive accuracy: predictions should be traceable to physically and chemically meaningful spectral regions to support validation, quality assurance, model refinement, and scientific discovery.

Given the recent introduction and growing interest in eXplainable Artificial Intelligence (XAI), a recent systematic review by Contreras and Bocklitz~\cite{contreras2025xai} confirmed that the field remains nascent within the spectroscopy domain, as the identified, earliest study directly mentioning both XAI and spectroscopy dates back to 2020. This is consequential because explainability serves several distinct purposes in AI. As discussed in the XAI literature, explanations may be required to \emph{justify} decisions, especially in settings demanding transparency and accountability; to \emph{control} systems by enabling auditing, debugging and detection of undesirable behavior; to \emph{improve} models through the identification of biases, brittle dependencies and suboptimal design choices; and to \emph{discover} novel patterns that can be turned into scientific or domain knowledge~\cite{adadi2018xai,vilone2021notions}. In spectroscopic modeling, these motivations highlight the need for explanations that are chemically grounded, able to distinguish genuine signals from artefactual or instrument-specific cues, and robust to the strong correlation structure of spectral variables. 

The XAI methods catalogued in the review~\cite{contreras2025xai} included perturbation-based approaches such as SHapley Additive exPlanations (SHAP)~\cite{lundberg2017shap} and Local Interpretable Model-agnostic Explanations (LIME)~\cite{ribeiro2016lime}, gradient-based attributions such as Class Activation Mapping (CAM/Grad-CAM)~\cite{selvaraju2017gradcam,zhou2016learning}, activation-map embeddings, and global surrogate models~\cite{contreras2025xai}. Among these, SHAP was the most prevalent technique. In parallel, several studies combining spectroscopy with traditional chemometrics techniques have been published over the years. They span a wide range of sample matrices and spectroscopic methods, including vis-NIR and mid-infrared for cocoa, coffee, and teff flour analysis \cite{santos2021nir,bona2017support,casarin2025determination}, XRF and GRS for evaluating soil fertility, soybean protein levels, and milk authentication \cite{ribeiro2025xrfgama,ribeiro2024optimization,de2023soybean,galvan2022low}, LIBS for assessing cadmium levels in rice roots and for compositional analysis of copper- and iron-based alloys \cite{wang2021quantitative,gupta2024compositional}, among many others. 

Although not always framed under the XAI umbrella, tools capable of providing insights into model behavior have also been leveraged, such as weighted coefficients in Multivariate Linear Models \cite{de2025development}, Variable Importance in Projection (VIP) scores in Partial Least Squares Regression (PLS) \cite{ribeiro2024optimization,ribeiro2025impact}, and Permutation Feature Importance (PFI) in Support Vector Machines (SVMs) \cite{da2022variable}. While useful, most of these explainability tools (both those traditionally applied in chemometrics and those grounded in XAI) have been borrowed from image analysis or general multivariate data settings, \textit{i.e.}, are not naturally designed to account for distinctive properties of spectroscopic measurements, such as physical continuity across adjacent variables, high correlation within spectral bands, and the direct correspondence between variables and physicochemical phenomena. Most methods attribute importance at the level of individual features (\textit{e.g.}, single wavenumbers or energy variables) rather than at the level of spectral zones, which can lead to fragmented and chemically inconsistent explanations~\cite{contreras2025xai}. Moreover, XAI techniques such as CAM and gradient-based attributions are restricted to differentiable architectures (\textit{e.g.}, Multi-Layer Perceptrons, MLPs, and Convolutional Neural Networks, CNNs), thereby excluding classical chemometric models (\textit{e.g.}, PLS and SVM) that remain widely used in spectral analysis.

On the other hand, interval-based variable selection methods such as Interval PLS (iPLS \cite{Norgaard2000}) and its synergy variant (siPLS \cite{shariati2010selection}) have long been used when combining spectroscopy and chemometrics to identify useful spectral windows by training independent PLS models on each candidate interval and selecting those that minimize prediction error. However, these methods are fundamentally ante-hoc, \textit{i.e.}, they operate during model construction to select features, not after training to explain a fixed model's decisions. As a result, they answer the question of which interval yields the most accurate predictive model, rather than which zones drove the reasoning of a model already deployed. 

Taken together, these aspects point to a scientific gap for explainability methods that are natively designed for spectral data, operate at the level of chemically meaningful spectral regions, communicate explanations directly aligned with the model's reasoning, remain readily interpretable by domain experts, and are model-independent and integrable with either ML or chemometric workflows.

To advance the field, we introduce Spectral Model eXplainer (SMX), a explainability framework developed for handling spectral-based models. It builds upon the Decision Predicate Graph (DPG) method introduced by Arrighi \textit{et al.} \citep{arrighi2024dpg}, which formalized the use of model-centric logical rules termed predicates (\textit{e.g.}, $Feature > Threshold$) and directed weighted graphs for explaining tree ensemble models. SMX adapts and substantially extends this foundation toward spectral-grounded outcomes. Specifically, it reconceptualizes explanation as a navigation problem across expert-informed spectral zones. Each zone is compressed into a representative PCA score, preserving the continuity and collinearity of spectral variables, from which predicates are extracted. Through a bagging strategy, multiple bags are generated, each proposing an ordered route through the defined spectral ranges. SMX then queries the underlying model through perturbations to identify the most relevant paths and integrates the results into a graph whose topology summarizes which predicates are repeatedly useful from the model’s perspective. 

This design yields practical advantages. SMX is structured as a model-agnostic framework, \textit{i.e.}, applicable to any spectral classifier, including PLS, SVM, and MLP. As predicates operate on aggregated spectral zones, explanations are derived considering the joint contribution of the multiple variables that compose each zone, being chemically plausible and readily communicable to domain experts. Furthermore, SMX belongs to the post-hoc XAI paradigm, providing explanations centered on the behavior of the trained models, rather than the features influencing their construction. Beyond identifying the most relevant spectral zones from the model's perspective and ranking their influence through graph-based centrality measures, each predicate carries an additional interpretive layer: its threshold-reconstructed spectral profile (called the \emph{threshold spectrum}) functions as a visual decision boundary that users can directly overlay on their measured spectra. This bridge between a statistical logical rule and a physicochemical spectral profile enables practitioners to visually assess how their samples relate to the model's learned boundaries, extracting intuitive and actionable insights that purely numerical importance scores extracted by existing explainability methods cannot provide. Moreover, the bagging strategy emphasizes relations that survive sampling disturbances, improving stability relative to single-instance perturbation methods. Finally, graph-theoretic metrics (\textit{e.g.}, Local Reaching Centrality, LRC) provide global explanations upon the captured model reasoning.

In this paper, SMX theory is formalized and its behavior for explaining spectral-based binary classifiers of different complexity (PLS, SVM, MLP) is assessed when handling synthetic and real spectral datasets. An extensive comparison with state-of-the-art XAI and chemometric methods (namely SHAP, PFI and VIP) is provided and a discussion of SMX limitations and directions for future studies is presented.

In summary, the main contributions are as follows:
\begin{itemize}
    \item The proposal of SMX, a post-hoc, global, and model-agnostic explainability framework tailored to the specific structural properties of spectral data.
    \item A zone-based predicate formulation that replaces isolated variable attribution with explanations grounded on meaningful spectral regions.
    \item The introduction of a spectral threshold reconstruction, enabling the mapping of logical predicates back to interpretable spectral profiles in the original measurement space.
\end{itemize}

\section{XAI-related concepts and terminologies}
\label{sec:xai_concepts}

Given the interdisciplinary nature of this study, this section briefly introduces the XAI concepts and evaluation criteria used throughout the analyses.  In our context, an \textit{explanation} is envisioned as a human-interpretable representation of the reasoning or behavior of a machine learning model~\cite{nauta2023anecdotal}, intended to help users understand, validate, and act upon its predictions. The XAI field addresses the challenge of generating such representations for complex, opaque predictors. Explanations may be \textit{local} (elucidating a single prediction) or \textit{global} (characterizing model behavior across all instances)~\cite{longo2024xai,speith2022review}. They may also be generated \textit{post-hoc}, after model training, or \textit{ante-hoc}, as an intrinsic property of the model architecture. This ante-hoc versus post-hoc distinction is consequential for method comparison: ante-hoc approaches such as interval-based PLS variants (iPLS, siPLS) optimize model construction by selecting spectral windows focusing on improving the models' performance, while post-hoc methods explain the behavior of models that have already been trained and deployed. Accordingly, post-hoc methods are typically model-agnostic, applicable to any trained predictor regardless of its internal structure~\cite{longo2024xai}.

Beyond these structural dimensions, recent literature has emphasized that explanation quality is inherently multi-dimensional and should be assessed against explicit criteria reflecting both the behavior of the underlying model and the needs of the intended users~\cite{vilone2021notions,nauta2023anecdotal}. In this study, particular attention is given to complementary properties that are especially relevant for spectral analysis: \textit{faithfulness}, \textit{stability}, \textit{composition}, \textit{simplicity}, and \textit{domain alignment}. Table~\ref{tab:xai_properties} summarizes each one in terms of general XAI concepts and translated to their meaning in spectroscopic domain. Furthermore, it is worth mentioning that the precise terminology used to refer to these properties varies across the XAI literature, where they are closely related to established notions of correctness, robustness, sensitivity, human-grounded interpretability, and application-grounded usefulness according to~\cite{vilone2021notions,nauta2023anecdotal,miller2019explanation}.

\begin{table*}[ht]
\centering
\caption{XAI quality properties evaluated and their 
         meaning in spectroscopic modeling contexts.}
\label{tab:xai_properties}
\resizebox{\linewidth}{!}{%
\small
\begin{tabular}{p{2.6cm} p{5.0cm} p{5.0cm}}
\toprule
\textbf{Property} & \textbf{General definition} & 
\textbf{Spectroscopic meaning} \\
\midrule
Faithfulness
& Degree to which the explanations reflect the model's actual decision process rather than merely 
  plausible post-rationalization~\cite{vilone2021notions,nauta2023anecdotal}.
& Important zones should genuinely drive model output, not merely 
  correspond to visually prominent peaks unrelated to the 
  classification. \\
\addlinespace
Stability
& Explanations remain consistent across repeated runs, minor input 
  perturbations, or modest data 
  variations~\cite{vilone2021notions,nauta2023anecdotal}.
& Small measurement noise, preprocessing choices, fluctuations in training data, or repeated runs should not produce substantially 
  different zone rankings. \\
\addlinespace
Composition
& Presentation format, organization, and structure of explanations aimed at favoring their clarity~\cite{nauta2023anecdotal}
& Delivering explanations through well-organized, structured, summarized spectral entities to facilitate user understanding and acting upon \\
\addlinespace
Simplicity
& Ability to reduce the number of entities involved in the explanation~\cite{vilone2021notions} 
& Explanations expressed as spectral zones or profiles in natural units 
  (\textit{e.g.}, keV, wavenumber) are more actionable than lists of hundreds of 
  individual variable scores. \\
\addlinespace
Domain alignment
& The extent to which explanations are well-aligned with domain knowledge~\cite{vilone2021notions,nauta2023anecdotal}.
& Outputs should correspond to known elemental signals or plausible spectral features rather than
arbitrary, noisy patterns lacking interpretive value \\
\bottomrule
\end{tabular} 
}
\end{table*}

\section{Spectral model explainer}
\label{sec:theory}

As mentioned, the SMX framework is conceptually inspired by the DPG \cite{arrighi2024dpg}, from which it inherits the predicate extraction, directed weighted graph structure, and LRC as a global importance metric. A detailed structural comparison between DPG and SMX summarizing similar components and novel extensions is provided in 
Appendix~\ref{appem:smx_dpg}. Specifically, DPG is designed for tree ensemble models operating on general tabular features, where each predicate corresponds directly to a decision node split extracted from the model internal logic, while SMX is agnostic and  motivated by the structure of spectral data. 

\begin{figure*}[htp!]
    \centering
    \includegraphics[width=\textwidth]{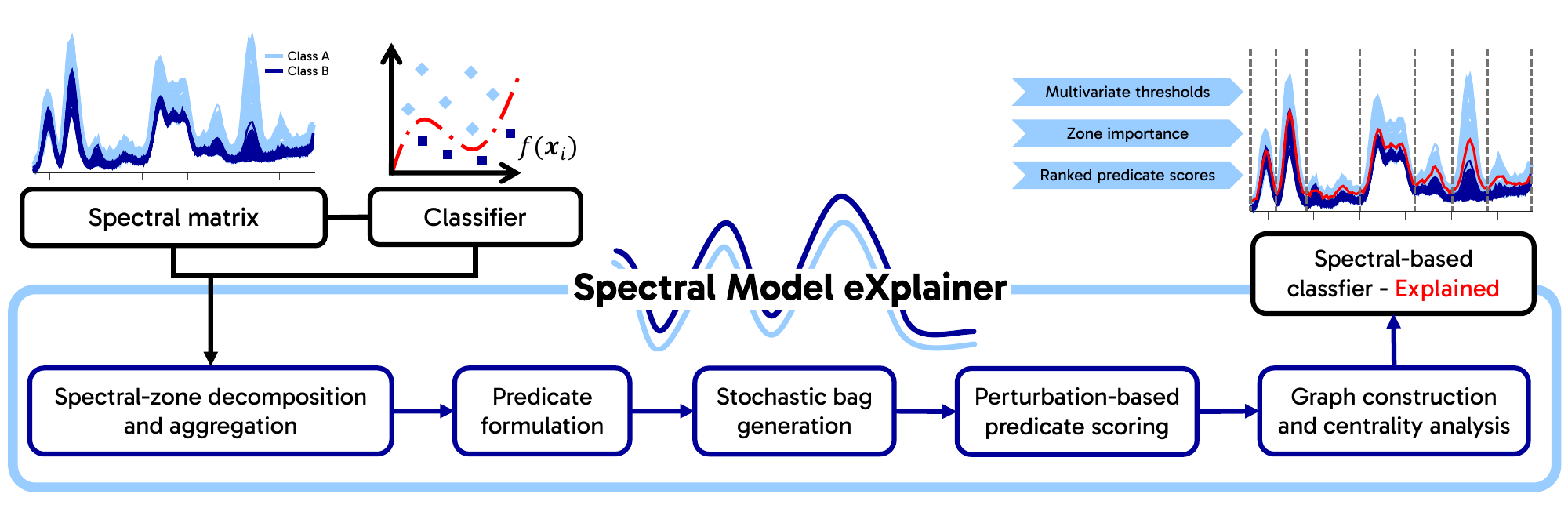}
    \caption{\small Pipeline of the Spectral Model eXplainer (SMX). Given a preprocessed spectral matrix and a trained supervised model, SMX extracts global post-hoc explanations.}
    \label{fig:smx_pipeline}
\end{figure*}

This section formalizes the SMX method. Let $\mathbf{X}\in\mathbb{R}^{n\times p}$ denote the preprocessed spectral matrix ($n$ samples, $p$ variables) and let $f$ be a trained supervised model whose predictions are $\hat{y}_i=f(\mathbf{x}_i)$. SMX proceeds through five main stages for extracting global, post-hoc explanations: (i)~spectral-zone decomposition and aggregation (Section \ref{sec:zones}), (ii)~predicate formulation (Section \ref{sec:predicates}), (iii)~stochastic bag generation (Section \ref{sec:bagging}), (iv)~perturbation-based predicate scoring (Section \ref{sec:perturbation}), and (v)~graph construction and centrality analysis (Section \ref{sec:graph}). Accordingly, the explanations delivered consist of a ranking of predicate relevance scores, from which it is possible to identify the most influential spectral features, the predicate intervals most associated with model's behavior, and the corresponding threshold spectra (Section \ref{sec:threshold_spectrum}). An overview is provided in the Figure \ref{fig:smx_pipeline}.

\subsection{Spectral-zone decomposition and aggregation}
\label{sec:zones}

The continuous spectrum is partitioned into $M$ meaningful zones in light of the studied dataset and the expert-informed spectral zones. Accordingly, each zone $Z_m$ is an index set defined by variable boundaries $\lambda_{\mathrm{start}^{(m)}}$ and $\lambda_{\mathrm{end}^{(m)}}$ (\textit{e.g.}, 6.0 $\le$ energy ranges $\le$ 6.8) so that:
\begin{equation}\label{eq:zone_def}
  Z_m = \bigl\{j : \lambda_{\mathrm{start}}^{(m)} \le \lambda_j \le \lambda_{\mathrm{end}}^{(m)}\bigr\},
  \qquad m=1,\dots,M,
\end{equation}
yielding $M$ sub-matrices $\mathbf{X}_{Z_m}\in\mathbb{R}^{n\times d_m}$, where $d_m=|Z_m|$.

Each zone is then summarized into a scalar per sample value via PCA with one principal component. Mathematically, each $Z_m$ is centered and the first principal-component loading vector $\mathbf{w}_1^{(m)}\in\mathbb{R}^{d_m}$ is obtained as the eigenvector associated with the largest eigenvalue of the sample covariance matrix:
\begin{equation}\label{eq:pca_loading}
  \mathbf{w}_1^{(m)}
  = \arg\max_{\|\mathbf{w}\|=1}
    \operatorname{Var}\!\bigl(\tilde{\mathbf{X}}_{Z_m}\,\mathbf{w}\bigr),
\end{equation}
where $\tilde{\mathbf{X}}_{Z_m}=\mathbf{X}_{Z_m}-\mathbf{1}_n\,\bar{\mathbf{x}}_{Z_m}^\top$ is the column-centered sub-matrix and $\bar{\mathbf{x}}_{Z_m}$ is the zone mean vector. The score of sample $i$ in zone $m$ is
\begin{equation}\label{eq:score}
  t_i^{(m)}
  = \bigl(\mathbf{x}_i^{(m)}-\bar{\mathbf{x}}_{Z_m}\bigr)^\top
    \mathbf{w}_1^{(m)}.
\end{equation}
This produces a score matrix $\mathbf{T}\in\mathbb{R}^{n\times M}$. The fraction of variance explained by PC1,
\begin{equation}\label{eq:varexp}
  \mathrm{VE}^{(m)}
  = \frac{\sigma_1^{2}}{\sum_{k=1}^{d_m}\sigma_k^{2}},
\end{equation}
quantifies how well the score captures the information content of the zone and is later used to weight edge importance in the graph.

Although approximating to one component, PCA aggregation offers two key advantages over simpler aggregators (\textit{e.g.}, sum, mean). First, it optimally weights each zone according to the direction of maximum variance, thereby providing a compact summary of the dominant variation within the zone. Second, the transformation is invertible: a scalar threshold $\tau$ can be projected back into the original spectral space as a multivariate \emph{threshold spectrum} (Section~\ref{sec:threshold_spectrum}), which is impossible with non-linear aggregators.

Algorithmically, the spectral-zone decomposition and PCA aggregation stage is detailed in the pseudocode of Algorithm~\ref{alg:smx_phase1} in the Appendix~\ref{app:pseudocode}).

\subsection{Predicate formulation}
\label{sec:predicates}

Predicates are logical rules derived from thresholding the zone-level scores. Each threshold induces a binary split of the samples, giving rise to two complementary rules: the samples above ($>$) or below-or-equal ($\le$) the limit.

Given a set of quantile levels chosen a priori by the practitioner $\mathcal{Q}=\{q_1,\dots,q_K\}$ (\textit{e.g.}, $\{0.2,0.4,0.6,0.8\}$), for each zone $m$ and each quantile $q_k$, the empirical quantile threshold $\tau_{m,k}=\mathcal{Q}_{q_k}(t_1^{(m)},\dots,t_n^{(m)})$ is computed and two complementary predicates are defined:
\begin{equation}\label{eq:predicates}
  P_{m,k}^{\le}\!:\; t_i^{(m)}\le\tau_{m,k},
  \qquad
  P_{m,k}^{>}\!:\; t_i^{(m)}>\tau_{m,k}
\end{equation}
The maximum number of predicates is $N_P=2MK$; duplicate predicates arising from tied quantile values are removed, yielding $N_P'\le N_P$ unique predicates. A binary indicator matrix $\mathbf{I}\in\{0,1\}^{n\times N_P'}$ records which samples satisfy each predicate.

Algorithmically, the predicate formulation stage is detailed in the pseudocode of the Algorithm~\ref{alg:smx_phase2} in the Appendix~\ref{app:pseudocode}).

\subsection{Stochastic bag generation}
\label{sec:bagging}


In high-dimensional spectral tasks, modest sample sizes, class imbalance, and outliers may lead to instability and bias when extracting the real importance of features through perturbation-based methods. Since SMX relies on perturbing the spectral zones (discussed in Section~\ref{sec:perturbation}), a stochastic bagging scheme is adopted to expose the predicate set to diverse but representative subsamples, reducing the variance of impact estimates and favoring predicates whose importance is recurrent across resamples.
Specifically, for each bag $b=1,\dots,B$, a subsample $\mathcal{S}_b\subset\{1,\dots,n\}$ of size $n_b$ (typically $0.8\,n$) is drawn without replacement. Within each bag, the samples satisfying a given predicate $P_j$ are collected as $\mathcal{S}_b^{(j)}=\{i\in\mathcal{S}_b:I_{i,j}=1\}$. Predicates for which $|\mathcal{S}_b^{(j)}|$ falls below a minimum-support threshold $n_{\min}$ of 20 \% of the training set size are excluded from that bag to ensure statistical reliability. The entire bagging, scoring, and graph-based pipeline is repeated independently across a predefined set of $R$ random seeds $\mathcal{R} = \{r_1, \ldots, r_R\}$, and the final centrality score of each predicate are averaged across $|\mathcal{R}|$ repetitions. This multi-seed averaging reduces the sensitivity of the final 
ranking to individual stochastic subsampling outcomes, acting as a variance-reduction mechanism. Accordingly, such strategy leads to hyperparameters to be defined a priori by the practitioner, namely the number of bags $B$, the subsample size $n_b$, and the $|\mathcal{R}|$ number of repetitions. 

\subsection{Perturbation-based predicate scoring}
\label{sec:perturbation}

The relevance of a predicate within each bag is quantified by measuring how much the model's output changes when the spectral information in the corresponding zone and subset of samples is perturbed. The underlying rationale is that if a zone is important to the model, replacing its values with an uninformative substitute should produce a noticeable shift in predictions.

Concretely, for a predicate $P_j$ associated with the $Z_m$ zone and the subset of samples $\mathcal{S}\equiv\mathcal{S}_b^{(j)}$, a perturbed input $\tilde{\mathbf{x}}_i$ is constructed by replacing only the variables belonging to $Z_m$ while leaving all other features intact:
\begin{equation}\label{eq:perturbation}
  \tilde{x}_{i,j}
  = \begin{cases}
      \operatorname{med}\{x_{k,j}:k=1,\dots,n\}, & j\in Z_m,\\[2pt]
      x_{i,j}, & j\notin Z_m,
    \end{cases}
\end{equation}
where $\operatorname{med}\{\cdot\}$ denotes the column-wise median computed over the entire training set. Even though other replacement statistics (mean, constant zero, min, max) are supported in the SMX method, the median is adopted as the default owing to its robustness to outliers.

The model is then queried on both the original and the perturbed instances, and an impact metric is computed. For models whose outputs are continuous numeric values (\textit{e.g.}, PLS), the default metric is the mean absolute error (MAE):
\begin{equation}\label{eq:mad}
  \operatorname{Imp}(P_j)
  = \frac{1}{|\mathcal{S}|}
    \sum_{i\in\mathcal{S}}
    \bigl|\hat{y}_i - \hat{y}_i^{\,\mathrm{pert}}\bigr|.
\end{equation}
For models whose primary output is a discrete class label rather than a continuous value (e.g., SVM and MLP), predicted class probabilities are used in place of raw predictions to obtain a continuously distributed impact measure. Specifically, the impact is quantified as the Probability Shift, defined as the mean absolute difference in predicted class probabilities before and after zone perturbation:
\begin{equation}\label{eq:imp_svm}
\mathrm{Imp}(P_j) = \frac{1}{|\mathcal{S}|}
    \sum_{i \in \mathcal{S}}
    \frac{1}{C}\sum_{c=1}^{C}
    \left| p_{i,c} - p_{i,c}^{\mathrm{pert}} \right|,
\end{equation}
where $p_{i,c}$ and $p_{i,c}^{\mathrm{pert}}$ are the predicted probabilities for class $c$ of sample $i$ before and after perturbation, respectively, and $C$ is the number of classes ($C = 2$ in the binary case). This formulation captures subtle perturbation effects on model confidence that may not manifest as changes in the predicted class label but nonetheless reflect a meaningful shift in the model's internal reasoning, providing a more granular and sensitive measure of zone influence than discrete classification accuracy. 

Furthermore, each predicate’s relevance score is normalized by the length of
the zone it represents to compensate for the tendency associated to wider zones,
which tend to produce larger perturbations simply due to their cumulative small
contributions across more variables (\textit{e.g.}, spectral background). Accordingly,
this normalization is achieved by dividing the raw impact score by the number
of variables in the zone ($d_m$), yielding a per-variable average score that allows
for a fair comparison across predicates regardless of their zone.
The corresponding bag generation and perturbation scoring processes are algorithmically detailed in the pseudocode of the Algorithm~\ref{alg:smx_phase3} in the Appendix~\ref{app:pseudocode}.

\subsection{Graph construction and centrality analysis}
\label{sec:graph}

The predicates and the perturbation scores computed for each bag are assembled into a global directed weighted graph $G=(V,E)$, in which $V$ denotes the set of predicates and $E$ the set of directed edges. Within each bag $b$, the predicates are sorted in descending order of perturbation impact and directed edges ($w$) are created between them:
\begin{equation}\label{eq:path}
  P_1^{(b)}
  \xrightarrow{w_1}
  P_2^{(b)}
  \xrightarrow{w_2}
  \cdots
  \xrightarrow{w_{L-1}}
  P_L^{(b)}
  \xrightarrow{w_L}
  \mathrm{Class}_{c^*},
\end{equation}
The weight of each edge is the perturbation score of its source node multiplied by the explained variance of the corresponding zone ($w_i^{\mathrm{VE}}=w_i\times\mathrm{VE}^{(m_i)}$). This causes zones where PC 1 captures a smaller share of the variance to have their weights reduced: a penalty for the quality of the representation. Additionally, $c^*$ is the majority predicted class among the samples satisfying the last predicate of the list. When paths from different bags produce the same edge $(u,v)$, the weights are accumulated by summation. After all bags have been processed, possible bidirectional edges are resolved by retaining only the direction with the larger cumulative weight to avoid cyclical routes that might impair the analysis of the graph`s centrality.

The global relevance of each predicate (or each node $v$ of $G$) is then quantified through its Local Reaching Centrality according to Mones \textit{et al.}~\cite{mones2012hierarchy}, which is the proportion of other nodes reachable from $v$ via directed paths, weighted by the average edge strength along those paths. Let $\pi(v,u)$ denote the shortest directed path from node $v$ to node $u$ in $G$, and let $|\pi(v,u)|$ denote the number of edges along that path. The LRC of node $v$ is computed as:
\begin{equation}\label{eq:lrc}
  \mathrm{LRC}(v)
  = \frac{1}{|V|-1}
    \sum_{\substack{u\in V\setminus\{v\}\\d(v,u)<\infty}}
    \frac{1}{|\pi(v,u)|}
    \sum_{k=1}^{|\pi(v,u)|} w_k^{(v,u)},
\end{equation}
where $w_k^{(v,u)}$ is the explained variance-weighted edge weight of the $k$-th edge along the shortest directed path $\pi(v,u)$ from $v$ to $u$, and the inner sum accumulates these weights over $|\pi(v,u)|$ edges composing that path. The outer sum runs over all nodes $u$ reachable from $v$ via at least one directed path ($d(v,u) < \infty$), normalised by the total number of other nodes $|V|-1$. Since predicates are ranked within each bag by perturbation impact, and directed edges encode this hierarchical ordering under the sampled data, LRC provides a global measure of predicate relevance. Predicates that repeatedly appear adjacent in high-impact positions across bags accumulate stronger edge weights and, consequently, higher LRC values. Accordingly, LRC summarises recurrent bag-wise evidence of influence in the ranking topology and can be interpreted as an importance-like score.

Importantly, the final graph does not encode direct causal or conditional dependencies between predicates but rather a principled aggregation of local oriented rankings prioritizing predicates based on their actual impact on the model’s behavior. Furthermore, as the bag generation is stochastic, the resulting graph topology and LRC scores might slightly vary across different random seeds. As mentioned in section \ref{sec:bagging}, the full SMX pipeline is repeated over multiple independent seeds $\mathcal{R}=\{r_1,\dots,r_R\}$, and the final centrality score of the $j$-th predicate is taken as the $j$-th mean $\overline{\mathrm{LRC}}(P_j) = {|\mathcal{R}|}^{-1}\sum_{s=1}^{R} \mathrm{LRC}_{r_s}(P_j)$. This multi-seed averaging acts as a variance-reduction mechanism analogous to ensemble averaging in which individual runs might fluctuate, but their mean converges to a robust centrality estimate as $|\mathcal{R}|$ grows. The graph construction and centrality analysis is summarized in Algorithm~\ref{alg:smx_phase5} (Appendix~\ref{app:pseudocode}).

%


\subsection{Threshold spectrum}
\label{sec:threshold_spectrum}

\begin{figure}[]
  \centering
  \includegraphics[width=0.7\linewidth]{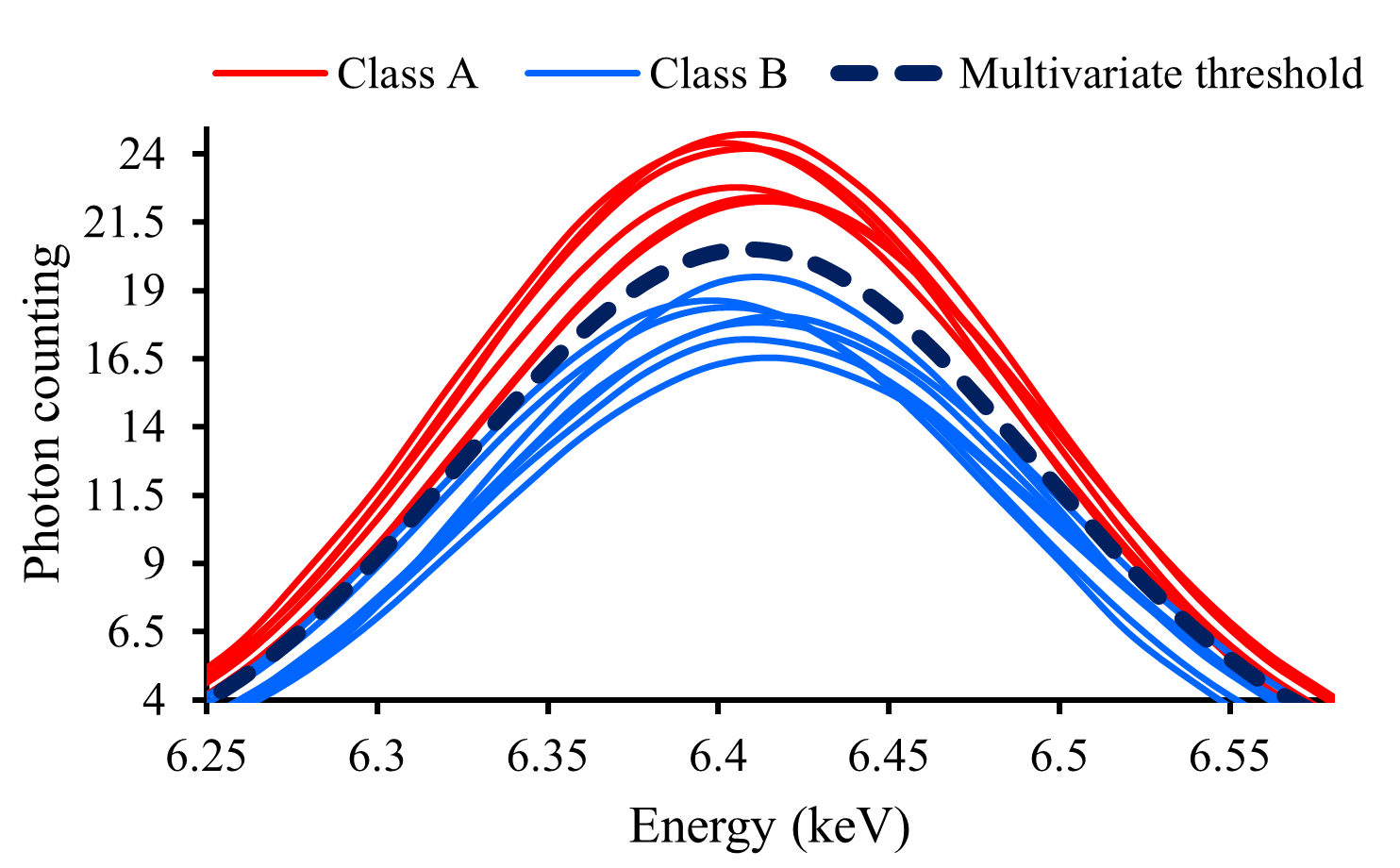}
  \caption{\small Illustrative example of a threshold spectrum corresponding to a predicate defined on the \textit{Fe K$\alpha$} zone from a XRF dataset covering two classes}
  \label{fig:threshold_spectrum}
\end{figure}

A distinctive feature of PCA-based aggregation is that the scalar threshold $\tau$ of a predicate can be mapped back to a full spectral profile. Because PCA is a linear projection, the inverse mapping yields the \emph{threshold spectrum}:
\begin{equation}\label{eq:threshold_spectrum}
  \boldsymbol{\tau}^{\,\mathrm{spectrum}}
  = \bar{\mathbf{x}}_{Z_m} + \tau\;\mathbf{w}_1^{(m)}
  \;\in\mathbb{R}^{d_m},
\end{equation}
which represents the spectral profile lying exactly on the decision boundary of the predicate within zone $Z_m$. Since the threshold and PCA are computed on the inputted data, the resulting spectrum is expressed in natural, inputted measurement units and can be directly overlaid on measured spectra. This capability transforms an abstract logical rule (\textit{e.g.}, ``Iron-connected zone $> 1.23$'') into a concrete model-related spectral boundary that domain experts can visually compare against their samples, bridging the gap between statistical explanations and physicochemical interpretation. Figure \ref{fig:threshold_spectrum} illustrates a typical example of this mapping.

The analytical interpretation of threshold spectra in a case study classification scenario and their practical utility for sample screening, instrument monitoring, and hypothesis generation, is fully discussed in sections \ref{sec:threshold_interpretation} and \ref{sec:case_study}.

\section{Materials and methods}
\label{sec:mat_and_methods}

\subsection{Datasets and preprocessing}
\label{sec:data}

This study assessed and compared SMX and baseline's explanations derived from both synthetic (for ablation purpose) and real spectral datasets covering binary classification tasks. Regression problems were not considered in this first proposal, since current XAI literature focuses predominantly on classification settings, and the extension to regression will be the subject of future studies.

The synthetic data were generated by modeling each spectrum as a superposition of Gaussian peaks plus additive noise. The appendix provides the detailed mathematical formulation of the data generation process (Section~\ref{subsec:synthetic_generation}). Specifically, the two-class scenario was constructed over a spectral axis of $p = 300$ points spanning the interval $[1, 600]$ (generic spectral units). The two classes shared common peaks centered at positions 150, 300, and 500, however with different amplitudes and widths for each class to introduce within-class variability. While the peaks at position 150 were highly more intense in Class $A$ than in Class $B$, the imbalance was less pronounced for the peaks at positions 300 and no imbalance was introduced for the peaks at position 500. This design created a well-defined ground truth where the spectral zone surrounding position 150 carry the strongest discriminative information, the zone around position 300 carries weaker but still relevant information, and the zone around position 500 is non-discriminative but differs from the random noise of the backgrounds. Table \ref{tab:synthetic_config} summarizes the parameters set. Accordingly, such a ground truth was leveraged to evaluate the sensitivity of SMX hyperparameters in the ablation study discussed further below.
 

\begin{table}[h]
  \caption{\small Configuration of the synthetic two-class spectral dataset.}
  \label{tab:synthetic_config}
  \resizebox{\linewidth}{!}{%
  \begin{tabular}{lccccccc}
    \toprule
    Class & $n$ & Peak center & $\bar{A}$ & $s_A$ & $\bar{\sigma}$ & $s_\sigma$ & $\sigma_\varepsilon$ \\
    \midrule
    \multirow{3}{*}{A} & \multirow{3}{*}{116} & 150 & 2.50 & 0.30 & 15.0 & 2.0 & \multirow{3}{*}{0.08} \\
      &   & 300 & 2.00 & 0.30 & 15.0 & 2.0 & \\
      &   & 500 & 0.50 & 0.30 & 15.0 & 2.0 & \\
    \midrule
    \multirow{3}{*}{B} & \multirow{3}{*}{126} & 150 & 0.10 & 0.005 & 15.0 & 2.0 & \multirow{3}{*}{0.10} \\
      &   & 300 & 0.80 & 0.30 & 14.0 & 1.5 & \\
      &   & 500 & 0.45 & 0.30 & 15.0 & 2.0 & \\
    \bottomrule
  \end{tabular}
  }
  \smallskip
  \scriptsize{$n$: number of samples; $\bar{A}$, $s_A$: mean and standard deviation of peak amplitudes; $\bar{\sigma}$, $s_\sigma$: mean and standard deviation of peak widths; $\sigma_\varepsilon$: baseline noise standard deviation. Common generation settings: $p=300$ points, spectral interval $[1,600]$,random seed $=42$.}
\end{table} 


Conversely, real spectral datasets based on XRF and GRS measurements were also employed to demonstrate SMX's behavior in real-world scenarios and to compare its outputs with other pertinent XAI methods. These data covered a wide range of applications and sample characteristics, including data extracted from soil, sediments, food, plant and bank-note instances. Eight datasets were considered in total, and Table~\ref{tab:real_datasets} summarizes their main details (further information is provided in the appendix~\ref{subsec:real_datasets})

\begin{table}[h]
  \caption{\small Evaluated binary-class XRF and GRS datasets}
  \label{tab:real_datasets}
  \resizebox{\linewidth}{!}{%
  \begin{tabular}{lccccccc}
    \toprule
    Dataset & $n$ & Class A & Class B & $p$ & Range & Zones & Domain \\
    \midrule
    Bank notes     & 407 & 251 & 156 & 785  & 2.74-22.71 & 15 & XRF \\
    Forage         & 195 &  58 & 137 & 971  & 1.40-20.81 & 22 & XRF \\
    Milk           & 383 & 143 & 240 & 781  & 2.66-22.62 & 10 & XRF \\
    Sediments      &  50 &  25 &  25 & 1166 & 1.40-13.05 & 19 & XRF \\
    Soil fertility & 212 & 110 & 102 & 590  & 1.32-13.10 & 20 & XRF \\
    Soil fertility &  80 &  56 &  24 & 516  & 95-610     &  9 & GRS \\
    Soil types & 156 &  77 &  79 & 374  & 57-430 &  9 & GRS \\
    Tomato         &  52 &  20 &  32 & 1049 & 2.12-23.08 & 17 & XRF \\
    \bottomrule
  \end{tabular}
  }
  \\
  \scriptsize{$n$: total number of samples; $p$: number of spectral variables; Zones: number of spectral zones; XRF ranges unit (keV); GRS ranges unit (number of channels)}
\end{table}

Standardized for ML modeling, all datasets were split into training/calibration (70 \% of samples) and test/validation (30 \% of samples) sets employing the Kennard-Stone (KS) algorithm~\cite{kennard1969computer} to ensure representativeness. The KS was applied individually to each class subset, and then the selected samples were merged to form the final training and validation sets so that they contain representative samples from each class. Afterwards, the data were preprocessed according to mean centering (synthetic data), Poisson scaling (\textit{i.e.}, scaling by the square root of the variables' mean) plus mean centering (XRF datasets), or Savitzky-Golay smoothing (window length 15, polynomial order 2) plus mean centering (GRS datasets). The preprocessing techniques were applied to the calibration sets and the resulting parameters were used to transform the test sets accordingly.

\subsection{Machine learning modeling}
\label{sec:ML}

For each dataset, three supervised classifiers were trained: Partial Least Squares discriminant analysis (PLS), Support Vector Machine (SVM), and Multilayer Perceptron (MLP). They were selected because they are widely used in chemometrics, machine learning, and spectroscopy domains, while also representing different levels of structural complexity and interpretability. PLS is a linear method commonly regarded as intrinsically more interpretable, particularly due to its available model-specific feature importance methods such as Variable Importance in Projection (VIP) scores. SVM with an RBF kernel is a non-linear learner that can capture complex relationships but is less interpretable, being grounded on maximizing the margin between classes in a transformed, high-dimensional space, which is not directly accessible to human interpretation. MLP is a feed-forward neural network that can model highly non-linear patterns but is often referred to as a black-box model due to its complex architecture and the distributed nature of its learned representations.

All models were implemented in Python using the \texttt{scikit-learn} library and specific configurations are summarized in Table~\ref{tab:ml_models}. For PLS, the number of latent variables was selected by 10-fold cross-validation on the training set, minimizing the root mean square error. For SVM and MLP, the remaining hyperparameters were kept at their default \texttt{scikit-learn} values unless otherwise stated. After training, all models were evaluated on the validation sets to confirm predictive performance before applying the explainability methods.

Specifically for MLP, a maximum of 10 training iterations was adopted deliberately, reflecting the relatively small sample sizes across the evaluated datasets. Given that the ratio of spectral variables to samples is high in most datasets considered, a shallow training budget acts as an implicit regularization mechanism, reducing the risk of overfitting to training data, and the use of an adaptive learning rate further supports efficient convergence within this budget. 

\begin{table}[ht]
  \centering
  \footnotesize
  \caption{\small Machine learning models and training configurations adopted in this study.}
  \label{tab:ml_models}
  \resizebox{\linewidth}{!}{%
  \begin{tabular}{p{1.3cm} p{2.0cm} p{6.5cm}}
    \toprule
    \textbf{Model} & \textbf{Implementation} & \textbf{Configuration} \\
    \midrule
    PLS
    & \texttt{PLSRegression}
    & Number of latent variables selected by 10-fold cross-validation on the training set, minimizing the root mean square error \\

    SVM
    & \texttt{SVC}
    & Radial Basis Function (RBF) kernel; remaining hyperparameters kept at default values \\

    MLP
    & \texttt{MLPClassifier}
    & Two hidden layers with 64 and 32 neurons; \texttt{tanh} activation; adaptive learning rate; maximum of 10 iterations; remaining hyperparameters kept at default values \\
    \bottomrule
  \end{tabular}
  }
\end{table}


\subsection{Ablation study}
\label{sec:ablation}

To uncover the SMX's sensitivity to its hyperparameters, an ablation study was conducted on the synthetic dataset. The hyperparameters evaluated included the quantile levels ($\mathcal{Q}$), the number of bags ($B$), the subsample size or number of samples per bag ($n_b$), and the number of internal repetitions ($|\mathcal{R}|$). Each hyperparameter was varied across a wide range of values while keeping the others fixed at their default settings. The range of values for each hyperparameter was chosen to cover both more conservative and more aggressive configurations. Table~\ref{tab:ablation_config} summarizes the specific values tested for each hyperparameter.

\begin{table}[ht]
  \footnotesize
  \caption{\small Hyperparameter range values tested in the ablation study.}
  \label{tab:ablation_config}
  \resizebox{\linewidth}{!}{%
  \begin{tabular}{lc}
    \toprule
    \textbf{Hyperparameter} & \textbf{Values tested} \\
    \midrule
    Quantile level steps ($\mathcal{Q}$)
    & $0.05, 0.1, 0.15, \textbf{0.2}, 0.25, 0.3$ \\
    Number of bags ($B$)
    & $5, \textbf{10}, 15, 20, 25, 30, 35, 40$ \\
    Subsample size ($n_b$)
    & $0.3n, 0.4n, 0.5n, 0.6n, 0.7n, \textbf{0.8n}, 0.9n$ \\
    Number of internal repetitions ($|\mathcal{R}|$)
    & $2, \textbf{4}, 6, 8, 10$ \\
    \bottomrule
  \end{tabular}
  }
  \scriptsize{SMX's default values are highlighted in bold. $n$ is the per-dataset number of samples and the Quantile level steps are defined as $\mathcal{Q}=\{q, 2q, \dots, (K-1)q\}$, where $q$ is the quantile level step and $K$ is the number of quantiles}
\end{table}

The resulting importance rankings were compared against the ground-truth discriminant zones to evaluate how changes in hyperparameters affect SMX's ability to correctly identify relevant spectral regions. As such, this analysis provided insights into the robustness of SMX-extracted explanations. 

\subsection{Explainability methods}
\label{sec:EXP_methods}

From each pair of trained ML model and dataset, SMX explanations were compared with established explainability methods to contextualize its behavior. 

SHAP~\cite{lundberg2017shap} is a well-established XAI method that attributes feature relevance through cooperative game theory, where each feature is a player and the model output is the payout. The Shapley value of feature $j$ for sample $\mathbf{x}_i$ is:
{\small
\begin{equation}\label{eq:shapley}
  \phi_j(\mathbf{x}_i)
  =\sum_{S \subseteq F \setminus \{j\}}
    \frac{|S|!\,(|F|-|S|-1)!}{|F|!}
    \bigl[f_{S\cup\{j\}}(\mathbf{x}_i) - f_{S}(\mathbf{x}_i)\bigr],
\end{equation}}
where $F$ is the feature set, $S$ spans all subsets excluding $j$, and $f_S$ is the expected model output conditioned on $S$. By construction, $\phi_j(\mathbf{x}_i)$ provides a local explanation. Global explanations are then extracted through the mean absolute Shapley values over all samples ($\bar{\phi}_j = \frac{1}{n}\sum_{i=1}^{n}|\phi_j(\mathbf{x}_i)|$). Because exact computation is exponential, KernelSHAP~\cite{lundberg2017shap} approximates Eq.~\eqref{eq:shapley} with a locally weighted linear surrogate on sampled coalition masks. In this study, SHAP explanations were computed with \texttt{KernelExplainer} (SHAP library), using raw model outputs for PLS and predicted class probabilities for SVM/MLP to reach a more granular quantification of feature importance, as opposed to discrete metrics such as accuracy. The time series-grounded variants of SHAP (\textit{e.g.}, Time-SHAP~\cite{bento2021timeshap}, GroupSegment-SHAP~\cite{kim2026groupsegment}) were not included as baseline because, although supporting window-based explanations, they are designed to handle temporal dependencies in sequential data, which is not the primary structure of spectral data. Spectral data, while ordered, do not exhibit the same kind of temporal dynamics, and therefore the assumptions underlying time series-specific methods are not well-suited for spectral analysis.


Permutation Feature Importance (PFI)~\cite{fisher2019all} is a method centered on quantifying feature relevance by the performance degradation induced by random permutation of each variable. Such tool is widely used in both the chemometrics and machine learning communities due to its simplicity and intuitive appeal. Formally:
\begin{equation}\label{eq:pfi}
  \operatorname{PFI}_j = \mathcal{L}(f, \mathbf{X}, \mathbf{y}) - \mathcal{L}(f, \mathbf{X}_{\mathrm{perm}(j)}, \mathbf{y}),
\end{equation}
where $\mathcal{L}$ is the performance metric, $f$ is the trained model, $\mathbf{X}$ and $\mathbf{y}$ are the original data, and $\mathbf{X}_{\mathrm{perm}(j)}$ is the dataset with feature $j$ permuted. Larger $\operatorname{PFI}_j$ indicates stronger dependence on feature $j$. Here, PFI was implemented using the average of 10 repetitions of random permutations and MAE computed on the raw predictions for PLS and on the predicted class probabilities for SVM/MLP, also for reaching more granular importance quantification.

As a traditional chemometric baseline for PLS, model-specific VIP scores~\cite{wold1993pls} were also extracted. The VIP score of variable $j$ is:
\begin{equation}\label{eq:vip}
  \mathrm{VIP}_j
  = \sqrt{p \sum_{l=1}^{LV} \frac{W_{j,l}^2\mathrm{SSY}_l}{\mathrm{SSY}_{\mathrm{total}} LV}},
\end{equation}
where $p$ is the number of variables, $LV$ is the number of latent variables, $\mathrm{SSY}_l$ is the response variance explained by latent variable $l$, $\mathrm{SSY}_{\mathrm{total}}$ is total response variance, and $W_{j,l}$ is the PLS weight of variable $j$ on latent variable $l$. VIP provides a global measure representing how much each variable contributes to explaining the variance in the response considering all model‘s latent variables. Its results were computed from the fitted PLS model parameters.

SMX-based explanations were generated adopting our implementation provided in the open-source \texttt{spectral-model-explainer} python library (freely available via the Python Package Index at the link \textcolor{blue}{\href{https://pypi.org/project/spectral-model-explainer/}{SMX}}\footnote{\url{https://github.com/joseviniciusr/SMX}}), setting the following default hyperparameters: $K=4$ quantiles ($\{0.2,0.4,0.6,0.8\}$), $B=10$ bags with subsample size $n_b=0.8\times$each training set size, and 4 repetitions for averaging. The personalized spectral zones were defined according to the expected chemical elements/signals in each dataset, based on prior knowledge of the physical principles of XRF and GRS (\textit{e.g.}, fluorescence/scattering features and radionuclide peaks \citep{Beckhoff2006, VanGrieken2001, Gilmore2008}), as well as the sample composition and measurement setup (\textit{e.g.}, excitation conditions, detector response and acquisition settings \citep{Knoll2010, VanGrieken2001}). Appendix~\ref{subsec:zone_details} summarizes the zone ranges for each dataset.

It is worth noting that ante-hoc variable selection methods such as iPLS and siPLS \citep{Norgaard2000,shariati2010selection} were not included as baselines in this comparison as they identify optimal spectral intervals by training and evaluating separate PLS models on each candidate window, and therefore operates at the model selection stage rather than the explanation stage. Since SMX, SHAP, PFI, and VIP all operate post-hoc on a single trained model, comparison with variable selection methods would conflate two methodologically distinct problems: predictive interval selection and post-hoc model explanation.

In light of the underlying principles of each explainability method, it is possible to qualitatively characterize their groundings according to the XAI evaluative dimensions (as clarified in section~\ref{sec:xai_concepts}) to contextualize the kind of evidence each one provide and under which assumptions the insights are comparable. Table~\ref{tab:XAI_adherence} summarizes such specificities. From this standpoint, SMX is referred to as a spectral-native explanation method, whose explanatory units and outputs are defined directly in terms of spectral structure.

\begin{table*}[h]
\centering
\caption{\small Explainability method based on game theory and its classification within the proposed taxonomy}
\label{tab:XAI_adherence}
\resizebox{\textwidth}{!}{%
\begin{tabular}{ccccc}
\hline
\textbf{} & \textbf{VIP} & \textbf{PFI} & \textbf{SHAP} & \textbf{SMX} \\ \hline
\textbf{Scope} & Global & Global & \begin{tabular}[c]{@{}c@{}}Local (native), \\ global (aggregated)\end{tabular} & \begin{tabular}[c]{@{}c@{}}Global \\ (subset-aware)\end{tabular} \\
\textbf{Timing} & Post-hoc & Post-hoc & Post-hoc & Post-hoc \\
\textbf{Agnosticism} & PLS-specific & Model-agnostic & Model-agnostic & Model-agnostic \\
\textbf{Faithfulness alignment} & \begin{tabular}[c]{@{}c@{}}Indirect \\ (latent-variance proxy)\end{tabular} & \begin{tabular}[c]{@{}c@{}}Direct (model's performance\\ degradation response)\end{tabular} & \begin{tabular}[c]{@{}c@{}}Direct (decomposes \\ model's prediction into \\ additive contributions)\end{tabular} & \begin{tabular}[c]{@{}c@{}}Direct (spectral \\ zone-wise \\ perturbation impact)\end{tabular} \\
\textbf{Stability nature} & Deterministic & Non-deterministic & Deterministic (as configured) & Non-deterministic \\
\textbf{Explanations simplicity} & Feature-level rankings & Feature-level rankings & \begin{tabular}[c]{@{}c@{}}Local aggregated \\ feature-level rankings\end{tabular} & \begin{tabular}[c]{@{}c@{}}Zone-based \\ predicate-level rankings\end{tabular} \\
\textbf{Output composition} & \begin{tabular}[c]{@{}c@{}}Dimensionless feature \\ importance scores\end{tabular} & \begin{tabular}[c]{@{}c@{}}Dimensionless feature \\ importance scores\end{tabular} & \begin{tabular}[c]{@{}c@{}}Aggregated feature marginal \\ contribution scores\end{tabular} & \begin{tabular}[c]{@{}c@{}}Zone-based predicates with \\ threshold-derived conditions \\ in natural spectral units\end{tabular} \\
\textbf{Grounding} & PLS theory & Permutation theory & Game theory & Spectral domain \\ \hline
\end{tabular}%
}
\end{table*}

\subsection{XAI evaluation metrics and statistical analysis}
\label{sec:XAI_metrics_and_stats}

Since SMX and compared baselines deliver outputs at different granularities and touch distinctly on the XAI dimensions (as described in Table~\ref{tab:XAI_adherence}), a comprehensive, quantitative evaluation was conducted to compare their explanations according to the discussed standard explainability desiderata (section \ref{sec:xai_concepts}), namely faithfulness, domain alignment, stability, and simplicity. 

For faithfulness, the explanations of each method were aggregated into ranked lists of spectral zones, extracted by sorting the feature-level importance scores in descending order, indexed by the zones they belong to, and dropping duplicates while keeping the first occurrences. Then, based on the incremental deletion strategy for assessing faithfulness~\cite{nauta2023anecdotal}, each method-specific zone ranking was used in a progressive top-$k$ masking strategy, where the corresponding spectral variables were sequentially masked (by replacing them with zeros) from the validation set, and the changes in the models' outcomes were quantified at each step using MAE (Eq.~\ref{eq:mad}) for PLS and Probability Shift (Eq.~\ref{eq:imp_svm}) for SVM and MLP. The area under the curve (AUC) of the resulted MAE/probability shifts vs $k$ curves were computed, so that higher AUC values indicate stronger faithfulness to the model's behavior. Then, the AUCs of SMX versus each of the other methods across all datasets were statistically compared via the pairwise Wilcoxon signed-rank tests~\cite{wilcoxon1945} to assess whether the observed per-model differences were statistically significant, with a significance level of $\alpha=0.05$. It is important to note that in each per-dataset comparison, the $k$ values for the compared methods were normalized to the same range (the minimum between the two methods' maximum $k$), since, in some cases, the SHAP values for some spectral zones were zero.

For domain alignment, the ranked zones generated for each dataset were compared with expert-defined plausibility lists that classified each spectral zone as either plausible (linked to known elemental signals or radionuclide peaks expected to carry discriminative information for the specific classification task and dataset) or non-plausible (corresponding to background regions, scattering continua, or energy ranges devoid of characteristic signals given the excitation conditions and sample matrices involved). Specifically, these classifications were established based on well-documented spectroscopic principles such as characteristic XRF lines\citep{Beckhoff2006, VanGrieken2001}, and GRS known radionuclide decay series and their characteristic emission energies \citep{Gilmore2008}. The complete plausibility classifications for each dataset are provided in Appendix~\ref{subsec:zone_details}. 


The rankings were then scored against the plausible set (discarding zones with zero importance scores across all involved spectral variables for that method), and cumulative agreements were computed by depth as
\begin{equation}\label{eq:agreement_rate_domain}
\mathrm{agreement\_rate}(k)=\frac{1}{k}\sum_{i=1}^{k}
\mathbb{I}\!\left[z_i\in\mathcal{Z}_{\mathrm{plausible}}\right],
\end{equation}
where $z_i$ is the zone at rank $i$ and $\mathbb{I}[\cdot]$ is the indicator function. Hence, $\mathrm{agreement\_rate}(k)$ is $1.0$ when all top-$k$ zones are plausible (best case) and $0.0$ when none is plausible (worst case). Analogously to the faithfulness evaluation, the area under the curve of the agreement rates across compatible $k$ values was extracted and the SMX results were statistically compared with the other methods via pairwise Wilcoxon signed-rank tests.

For simplicity, the importance scores within each explainer (predicates for SMX and individual variables for SHAP, PFI, and VIP) were normalized so that they sum to 1.0, \textit{i.e.}, dividing each variable's importance by the total importance across all variables for that method. The lower the number of features cumulating a high proportion of the total importance (\textit{e.g.}, 80 \%) the simpler the explanation. Because the number of features greatly differs across methods (\textit{e.g.}, SHAP and PFI provide importance scores for all variables, while SMX provides importance scores for predicates), the cumulative importance of the top-20 features was computed for each method, model and dataset, as it represented a plateau in the cumulative importance curves for most cases. The AUC of the resulted curves were then statistically compared between SMX and the other methods across models and datasets via pairwise Wilcoxon signed-rank tests. Conversely, the explanations' composition was also qualitatively evaluated through detailed discussions on the format, organization and structure of the explanations provided by each method.

For stability, only the non-deterministic methods (SMX and PFI) were compared. For each model-dataset pair, the explainers were rerun on the same setting but with 10 different internal seeds, producing 10 explainability lists. Then, the lists of all possible seed pairs were compared via Rank-Biased Overlap (RBO)~\cite{webber2010rbo}, yielding $\binom{10}{2}=45$ pairwise scores per method, model, and dataset. RBO is an intersection-based similarity method specifically designed for quantitatively comparing ranked lists, providing a score in $[0,1]$ where 0 indicates completely disjoint rankings and 1 indicates identical rankings (more details can be found in~\cite{webber2010rbo}). Accordingly, for each seed pair $(r_a,r_b)$, the instability score 1-$\mathrm{RBO}_{ab}$ was computed (with top-weighting parameter $\rho=0.7$), so that the lower the values, the higher the reproducibility of the explainer under internal stochastic variation. Given each dataset, the median instability score across all seed pairs was computed and the per-model SMX and PFI results were statistically compared via pairwise Wilcoxon signed-rank tests.

\section{Results and discussion}
\label{sec:results}

\subsection{Spectral datasets and modeling results}
\label{sec:results_spectral}

\begin{figure}[]
  \centering
  \includegraphics[width=1\linewidth]{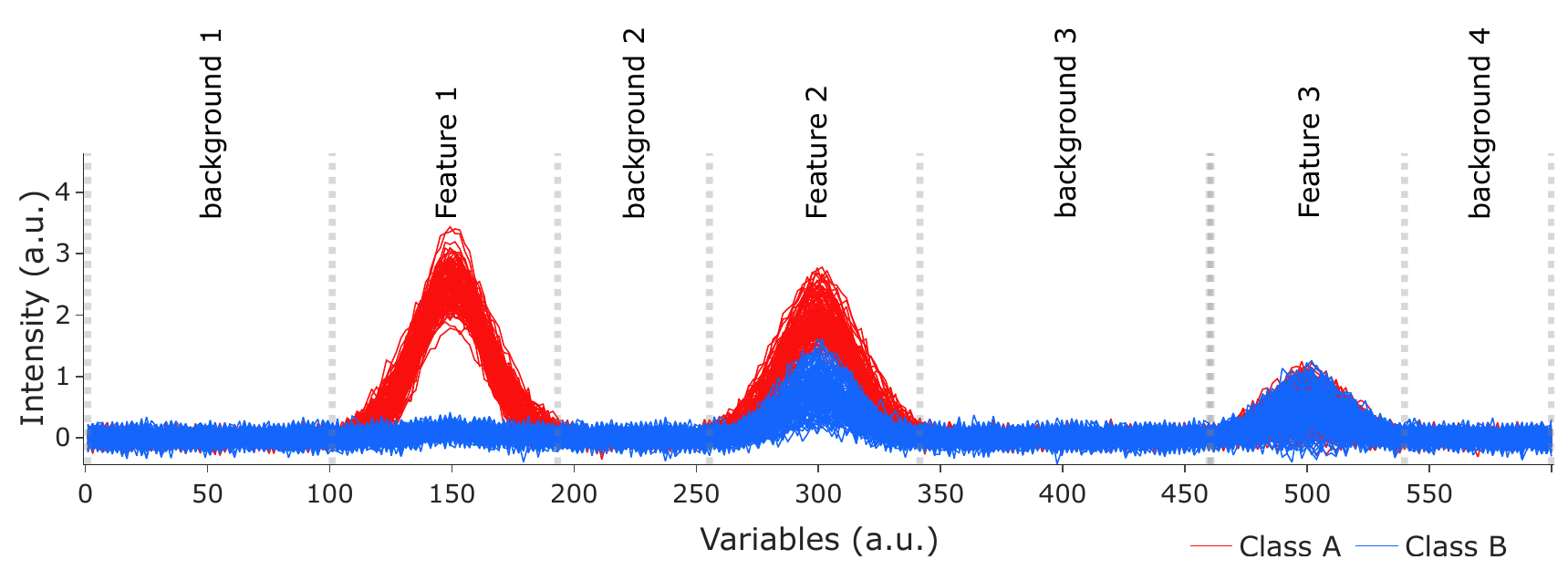}
  \caption{\small Synthetic-generated spectra, colored by class. Vertical delimited areas indicate the defined spectral zones. a.u.: arbitrary units}
  \label{fig:synthetic_spectra}
\end{figure}

\begin{figure*}[h]
  \centering
  \includegraphics[width=1\textwidth]{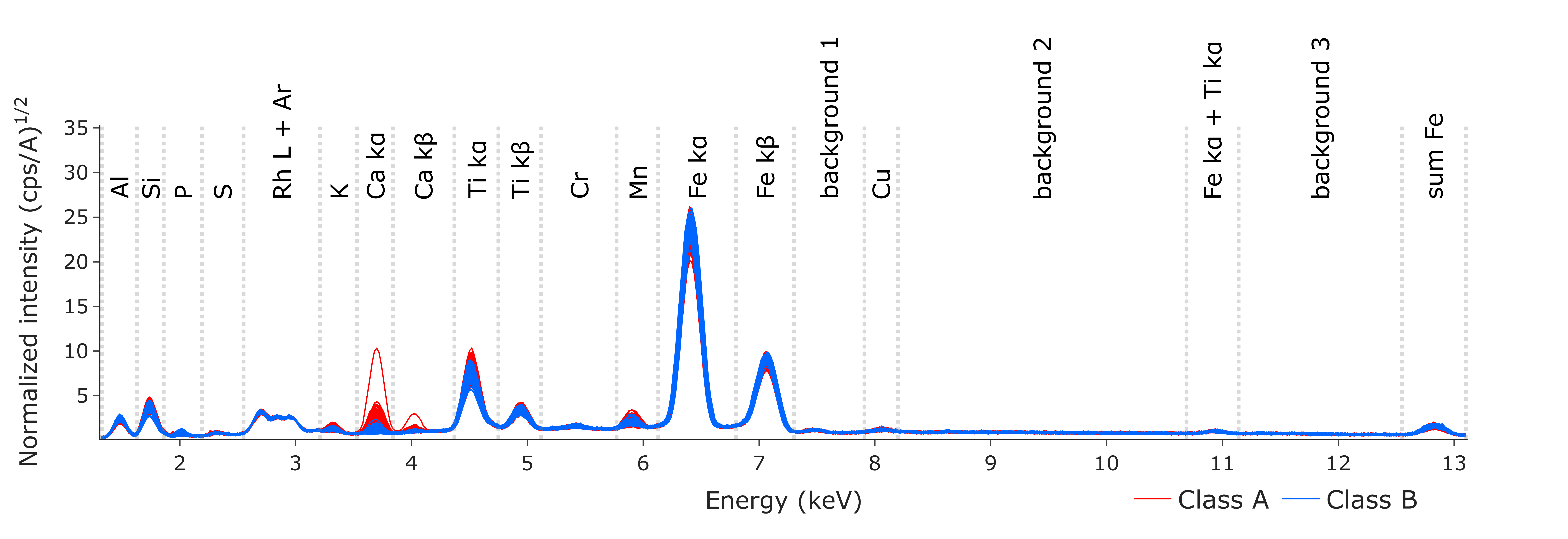}
  \caption{Poisson preprocessed spectra of the soil dataset, colored by class. Vertical delimited areas indicate the defined spectral zones. Class A corresponds to eutrophic soils, while class B corresponds to dystrophic soils}
  \label{fig:xrf_spectra}
\end{figure*}

Figures \ref{fig:synthetic_spectra} and \ref{fig:xrf_spectra} show the synthetic and a preprocessed XRF spectra (of the soil fertility dataset) to serve as a representative examples and to visually contextualize the defined spectral zones. The synthetic data (Figure \ref{fig:synthetic_spectra}) exhibit the expected Gaussian peaks at the predefined positions, with variability in their amplitudes and widths across samples, as well as a noisy baseline. The discriminant peaks at positions 150 and 300 presented different levels of class separation, with the former being more intense and well-separated between classes than the latter, while the shared peak at positions 500 is present in both classes without clear separation. The XRF spectra (Figure \ref{fig:xrf_spectra}) display multiple peaks corresponding to different soil-related elemental signals, with varying intensities across samples and classes.

Table~\ref{tab:ml_performance} summarizes the predictive performance of the trained ML models. Overall, the learners achieved high accuracy on both training and test sets, validating the adopted modeling configurations and indicating that the relevant discriminative structure was effectively captured within the allotted training procedure. This trend was particularly evident for the Bank notes, Forage, Soil types, and Synthetic datasets, in which the derived accuracy on the test set ranged from 0.98 to 1.0 across models. 

Nevertheless, some exceptions were observed for more challenging datasets, notably Soil (GRS), Sediments, Tomato, and Milk. It is noteworthy that MLP models generally underperformed relative to PLS and SVM, with more pronounced differences, especially for such datasets. Such outcomes are consistent with dataset-level complexity, namely class imbalance, small sample sizes, and high spectral dimensionality, rather than with premature training termination, since these same datasets also challenged PLS and SVM models.

\begin{table}[]
  \scriptsize
  \setlength{\tabcolsep}{7pt}
  \caption{\small Predictive performance of trained ML models across datasets}
  \label{tab:ml_performance}
  \resizebox{\linewidth}{!}{%
  \begin{tabular}{llcccccc}
    \toprule
    \textbf{Model} & \textbf{Dataset} & \multicolumn{3}{c}{\textbf{Train}} & \multicolumn{3}{c}{\textbf{Test}} \\
    \cmidrule(lr){3-5} \cmidrule(lr){6-8}
    & & \textbf{Acc} & \textbf{Sen} & \textbf{Spec} & \textbf{Acc} & \textbf{Sen} & \textbf{Spec} \\
    \midrule
    PLS &   & 0.99 & 0.98 & 1.00 & 0.99 & 0.98 & 1.00 \\
    SVM & Bank notes & 0.99 & 0.98 & 1.00 & 1.00 & 1.00 & 1.00 \\
    MLP &   & 1.00 & 1.00 & 1.00 & 1.00 & 1.00 & 1.00 \\
    \midrule
    PLS &   & 1.00 & 1.00 & 1.00 & 1.00 & 1.00 & 1.00 \\
    SVM & Forage & 1.00 & 1.00 & 1.00 & 1.00 & 1.00 & 1.00 \\
    MLP &   & 0.99 & 0.98 & 1.00 & 1.00 & 1.00 & 1.00 \\
    \midrule
    PLS &   & 0.82 & 0.88 & 0.73 & 0.77 & 0.83 & 0.65 \\
    SVM & Milk & 0.97 & 0.99 & 0.93 & 0.84 & 0.85 & 0.84 \\
    MLP &   & 0.98 & 0.97 & 1.00 & 0.72 & 0.63 & 0.88 \\
    \midrule
    PLS &  & 1.00 & 1.00 & 1.00 & 1.00 & 1.00 & 1.00 \\
    SVM & Sediments & 0.79 & 0.82 & 0.76 & 0.69 & 0.50 & 0.88 \\
    MLP* &  & 0.74 & 0.88 & 0.59 & 0.50 & 1.00 & 0.00 \\
    \midrule
    PLS &   & 0.88 & 0.89 & 0.87 & 0.84 & 0.90 & 0.79 \\
    SVM & Soil (XRF) & 0.93 & 0.94 & 0.91 & 0.88 & 0.97 & 0.79 \\
    MLP &   & 0.82 & 0.86 & 0.79 & 0.77 & 0.84 & 0.70 \\
    \midrule
    PLS &   & 0.73 & 0.12 & 0.97 & 0.72 & 0.12 & 1.00 \\
    SVM & Soil (GRS) & 0.75 & 0.12 & 1.00 & 0.72 & 0.12 & 1.00 \\
    MLP* &   & 1.00 & 1.00 & 1.00 & 0.44 & 0.63 & 0.35 \\
    \midrule
    PLS &   & 1.00 & 1.00 & 1.00 & 1.00 & 1.00 & 1.00 \\
    SVM & Soil types & 1.00 & 1.00 & 1.00 & 1.00 & 1.00 & 1.00 \\
    MLP &   & 1.00 & 1.00 & 1.00 & 1.00 & 1.00 & 1.00 \\
    \midrule
    PLS &   & 1.00 & 1.00 & 1.00 & 1.00 & 1.00 & 1.00 \\
    SVM & Synthetic & 1.00 & 1.00 & 1.00 & 1.00 & 1.00 & 1.00 \\
    MLP &   & 1.00 & 1.00 & 1.00 & 1.00 & 1.00 & 1.00 \\
    \midrule
    PLS &   & 0.75 & 0.82 & 0.64 & 0.69 & 0.80 & 0.50 \\
    SVM & Tomato & 0.86 & 1.00 & 0.64 & 0.63 & 0.70 & 0.50 \\
    MLP &   & 0.78 & 0.82 & 0.71 & 0.62 & 0.90 & 0.17 \\
    \bottomrule
  \end{tabular}%
  }
  \footnotesize{Acc: accuracy; Sen: sensitivity; Spec: specificity. Models marked * achieved test accuracy at or below the full samples' majority-class baseline}
\end{table}

 Two model–dataset combinations produced test accuracy at or below the full samples' majority-class baseline: Soil (GRS) MLP (acc = 0.44, majority-class baseline = 0.70) and Sediments MLP (acc = 0.50, majority-class baseline = 0.50). Additionally, the Soil (GRS) SVM model achieved high overall accuracy (0.72) but near-zero sensitivity (0.12), indicating a degenerate classifier dominated by majority-class predictions. In these cases, the underlying classifiers did not capture a reliable discriminative structure in the spectral data, and the derived explanations tend to confuse the importance of features related to noisy background regions with genuine element-based signals, regardless of the explainability method applied. 

This finding itself carries a methodological implication: explainability methods cannot substitute for predictive performance. A high-ranking zone from a near-random model provides no actionable analytical insight. Practitioners should always confirm model reliability before interpreting zone importance scores, and explainability pipelines should incorporate a performance gate as a prerequisite for explanation generation.


\subsection{Explainability evaluation analysis}
\label{sec:xai_evaluation}

According to the XAI evaluative dimensions (faithfulness, domain alignment, stability and simplicity) the comparisons between SMX and baselines are summarized as follows.  

\subsubsection{Faithfulness}

Regarding faithfulness, the mean results across datasets for the first 9 most important zones are shown in Figure~\ref{fig:faithfulness} (dataset-specific results are provided in the supplementary material). The 9th zone was chosen as the maximum depth for visualization purposes and general interpretability, since it is the highest common depth across all datasets. However, the statistical comparisons were performed at the dataset level using the maximum compatible depth for each method pair, as described in section~\ref{sec:XAI_metrics_and_stats}. The progressive top-$k$ masking curves consistently increased prediction changes for all methods, indicating that their top-ranked zones were genuinely the most influential to model behavior. Although differences among methods were small, VIP for PLS showed a slightly weaker effect in the first three zones, resulting in a lower AUC throughout the zones. To complement this analysis, paired Wilcoxon signed-rank tests across datasets, focusing on comparisons involving SMX, are reported in Table~\ref{tab:wilcoxon_faithfulness_smx}.

\begin{figure}[h]
  \centering
  \includegraphics[width=1.0\linewidth]{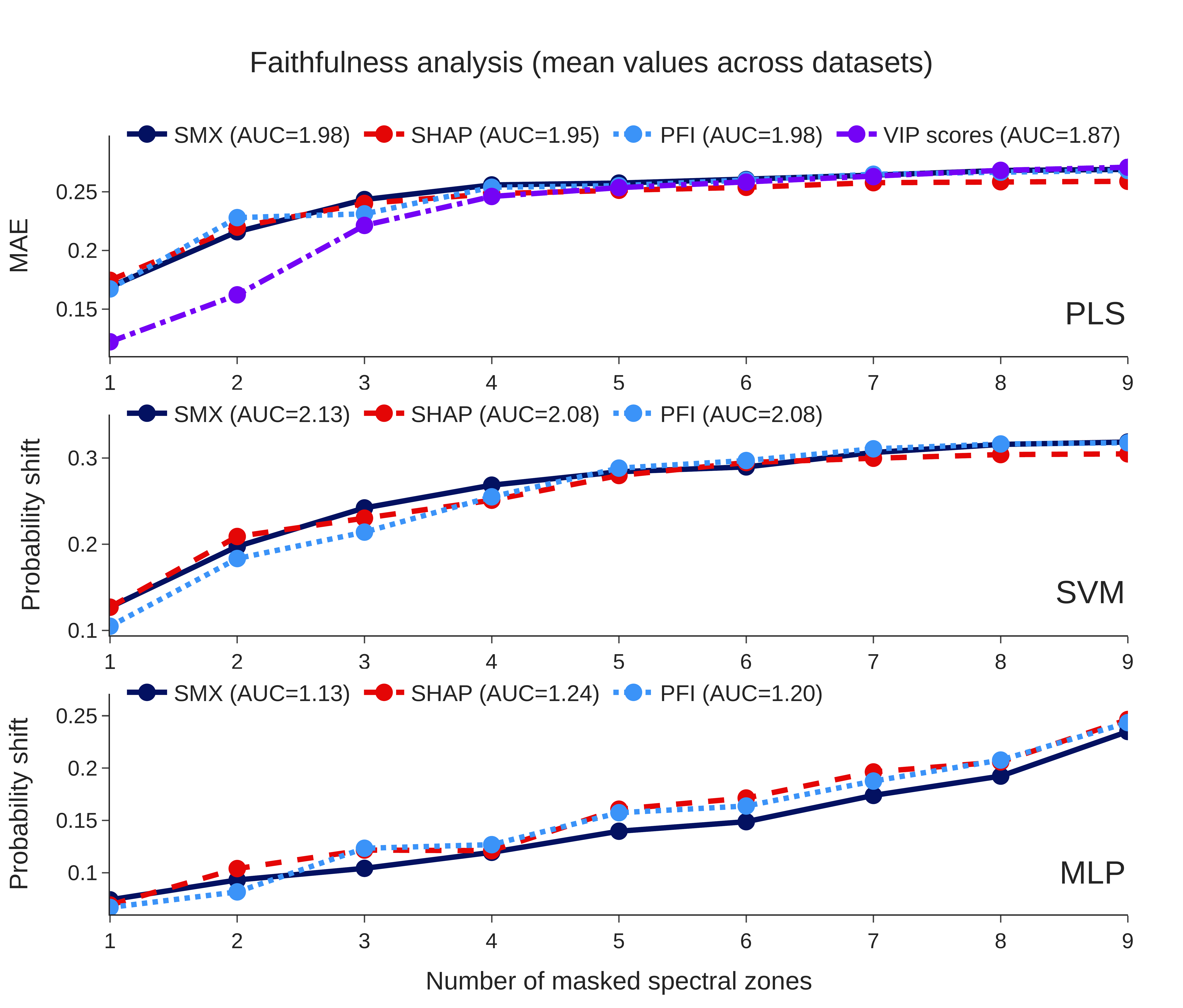}
  \caption{\footnotesize Faithfulness analysis via progressive top-$k$ masking. The plots show the per-$k$ mean MAE (for PLS) and probability shift (for SVM and MLP) across the datasets (\textit{n=}8). The legends indicate the mean area under the curve (AUC) encompassing the results for the first 9 most important zones for each method. The 9th zone was chosen as the maximum depth for visualization purposes, since it is the highest common depth across all datasets. However, the statistical comparisons were performed at the dataset level using the maximum compatible depth for each method pair, as described in section~\ref{sec:XAI_metrics_and_stats}}
  \label{fig:faithfulness}
\end{figure}

\begin{table}[]
  \centering
  \scriptsize
  \setlength{\tabcolsep}{6pt}
  \caption{Wilcoxon signed-rank results for faithfulness, focusing on comparisons involving SMX across datasets (\textit{n=}8). \textbf{Median diff} corresponds to the paired median of SHAP, PFI or VIP $-$ SMX}
  \label{tab:wilcoxon_faithfulness_smx}
  \begin{tabular}{llllr}
    \toprule
    \textbf{Model} & \textbf{Comparison} & \textbf{Median diff} & \textbf{p-value} \\
    \midrule
    PLS & PFI$-$SMX & -0.006 & 0.742 \\
    PLS & SHAP$-$SMX & -0.039 & 0.057 \\
    PLS & VIP$-$SMX & 0.001 & 0.844 \\
    SVM & PFI$-$SMX & 0.010 & 0.999 \\
    SVM & SHAP$-$SMX & -0.079 & 0.195 \\
    MLP & PFI$-$SMX & 0.053 & 0.078 \\
    MLP & SHAP$-$SMX & 0.102 & 0.009 \\
    \bottomrule
  \end{tabular}
\end{table}

Overall, the Wilcoxon results supported non-detectable faithfulness differences (considering the sampling size, $n=8$) between SMX and the baselines in six out of seven comparisons (Table~\ref{tab:wilcoxon_faithfulness_smx}). The only significant difference was observed for MLP in the SHAP vs SMX comparison ($p-value=0.009$), indicating stronger faithfulness scores for SHAP under that specific model setting. This outcome suggest that, in general, SMX is competitive with the baselines in terms of faithfulness.

\subsubsection{Domain alignment}

\begin{figure}[]
  \centering
  \includegraphics[width=1.0\linewidth]{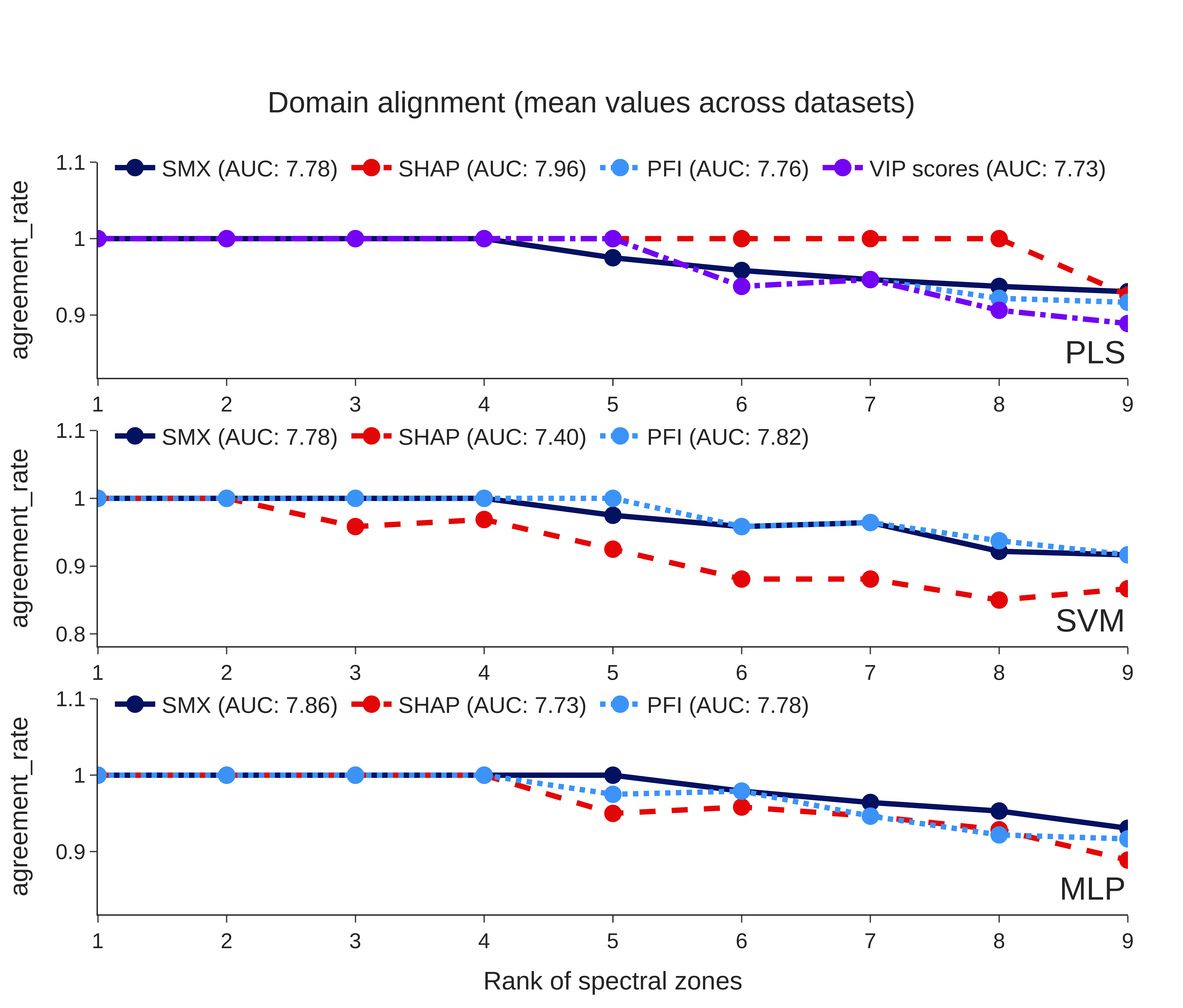}
  \caption{\footnotesize Domain alignment analysis via agreement with expert-defined plausible zones. The plots show the per-zone mean cumulative agreement rate across the datasets (\textit{n=}8) for each method. The legends indicate the mean area under the curve (AUC) encompassing the results for the first 9 most important zones for each method. The 9th zone was chosen as the maximum depth for visualization purposes, since it is the highest common depth across all datasets. However, the statistical comparisons were performed at the dataset level using the maximum compatible depth for each method pair, as described in section~\ref{sec:XAI_metrics_and_stats}}
  \label{fig:domaind_alignment}
\end{figure}

For domain alignment, Figure~\ref{fig:domaind_alignment} reports mean cumulative agreement rates for the first 9 most important zones across datasets (per-dataset detailed domain alignment results are available in supplementary materials). Similarly to faitfulness analyses, the 9th zone was chosen as the maximum depth for general interpretability and the statistical comparisons were performed at the dataset level using the maximum compatible depths. The explainers showed similar trends, with comparable AUCs. SHAP was slightly higher for PLS, whereas SMX and PFI were higher for SVM and MLP. Paired Wilcoxon comparisons (Table~\ref{tab:wilcoxon_domain_smx}) supported this pattern. For PLS, all SMX-centered comparison were non-significant, exhibiting non-detectable difference given the sampling size. For SVM and MLP, SHAP$-$SMX showed significant negative median differences ($p-value=0.046$ and $p-value=0.016$, respectively), favoring SMX. In MLP, PFI$-$SMX was also significant ($p-value=0.023$), again favoring SMX. Hence, these inferential results indicate that SMX is competitive and, in specific scenarios, statistically superior in domain alignment (within the sampling size context).

\begin{table}[]
  \centering
  \scriptsize
  \setlength{\tabcolsep}{6pt}
  \caption{Wilcoxon signed-rank results for domain alignment, focusing on comparisons involving SMX across datasets (\textit{n=}8). \textbf{Median diff} corresponds to the paired median of SHAP, PFI or VIP $-$ SMX}
  \label{tab:wilcoxon_domain_smx}
  \begin{tabular}{llllr}
    \toprule
    \textbf{Model} & \textbf{Comparison} & \textbf{Median diff} & \textbf{p-value} \\
    \midrule
    PLS & PFI$-$SMX & 0.036 & 0.999 \\
    PLS & SHAP$-$SMX & 0.000 & 0.500 \\
    PLS & VIP$-$SMX & 0.002 & 0.844 \\
    SVM & PFI$-$SMX & 0.056 & 0.999 \\
    SVM & SHAP$-$SMX & -0.420 & 0.046 \\
    MLP & PFI$-$SMX & -0.276 & 0.023 \\
    MLP & SHAP$-$SMX & -0.421 & 0.016 \\
    \bottomrule
  \end{tabular}
\end{table}

Despite that, a structural asymmetry should be acknowledged in this comparison: SMX operates natively on pre-defined zones and therefore can not select variables outside these boundaries, whereas SHAP, PFI, and VIP operate at the variable level and were aggregated at post-hoc timing. This difference means that SMX's domain alignment is partly a function of zone quality rather than purely of explanation quality. Nevertheless, the fact that variable-level methods can select variables within non-plausible background regions, as reflected in their lower alignment scores for SVM and MLP, supports the practical value of zone-constrained operation.

\subsubsection{Instability}

\begin{figure}[]
  \centering
  \includegraphics[width=1.0\linewidth]{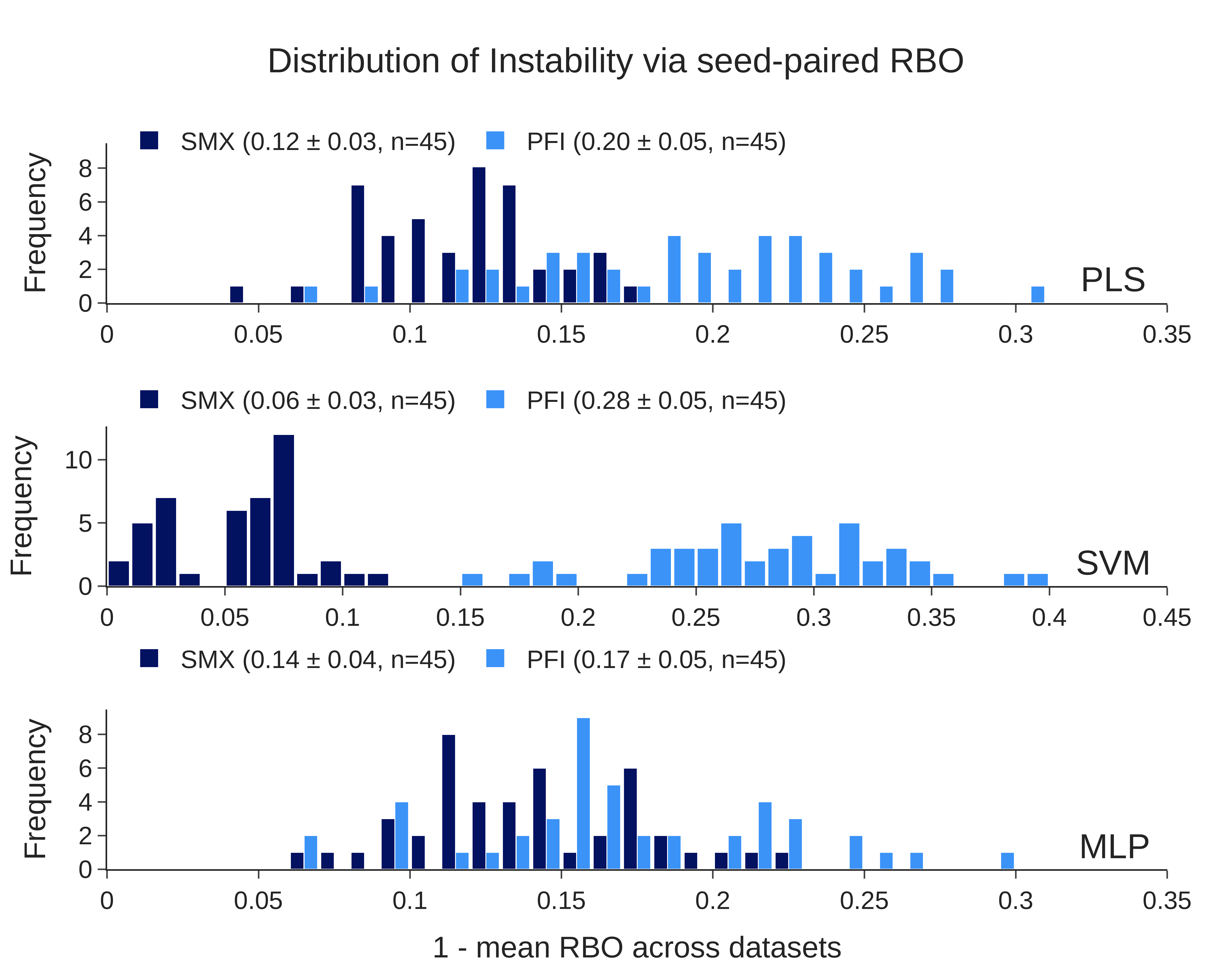}
  \caption{\small Stability analysis via pairwise RBO comparisons between the ranked predicate (SMX) and variable importance (PFI) lists obtained from 10 different random seeds for SMX and PFI, both with 4 internal repetitions. The histograms show the distribution of instability scores (1-RBO) across all seed pairs (\textit{n=}45) averaged across the datasets (\textit{n=}8) for each method. The mean $\pm$ standard deviation of the instability scores are indicated in the legends. The lower the instability score, the higher the stability of the method under internal stochastic variation}
  \label{fig:stability}
\end{figure}

For instability, SMX and PFI were compared through histograms of instability scores (1-RBO) over all 45 seed pairs possible combinations. The results (Figure~\ref{fig:stability}) are reported as mean pairwise values across datasets (individual histograms are presented in the supplementary materials) for overall interpretability, but the statistical comparisons were performed at the dataset level. At the raw-output level (predicate lists for SMX and variable lists for PFI), SMX generally showed lower instability, especially for SVM. Wilcoxon tests (Table~\ref{tab:wilcoxon_stability_smx}) showed non-significant differences for PLS and MLP given the sampling size, and a significant difference for SVM favoring SMX (median difference $-0.305$ and $p-value=0.016$). A plausible explanation is the SMX's ensemble design, which reduces sensitivity to random fluctuations during explanation generation. Given that PFI is a widely used non-deterministic baseline in chemometrics and machine learning, these results reinforce SMX as a promising, reproducible alternative for explaining spectral-based models. SHAP and VIP were not included in the stability analysis since they are deterministic.


\begin{table}[]
  \centering
  \scriptsize
  \setlength{\tabcolsep}{6pt}
  \caption{\small Wilcoxon signed-rank results for instability (1-RBO), SMX vs PFI across datasets (\textit{n=}8). \textbf{Median diff} corresponds to the paired median of SHAP, PFI or VIP $-$ SMX}
  \label{tab:wilcoxon_stability_smx}
  \begin{tabular}{llllr}
    \toprule
    \textbf{Model} & \textbf{Comparison} & \textbf{Median diff} & \textbf{p-value} \\
    \midrule
    PLS & PFI$-$SMX & -0.096 & 0.195 \\
    SVM & PFI$-$SMX & -0.305 & 0.016 \\
    MLP & PFI$-$SMX & -0.104 & 0.313 \\
    \bottomrule
  \end{tabular}
\end{table}


\subsubsection{Simplicity and compositional quality}

\begin{figure}[]
  \centering
  \includegraphics[width=1.0\linewidth]{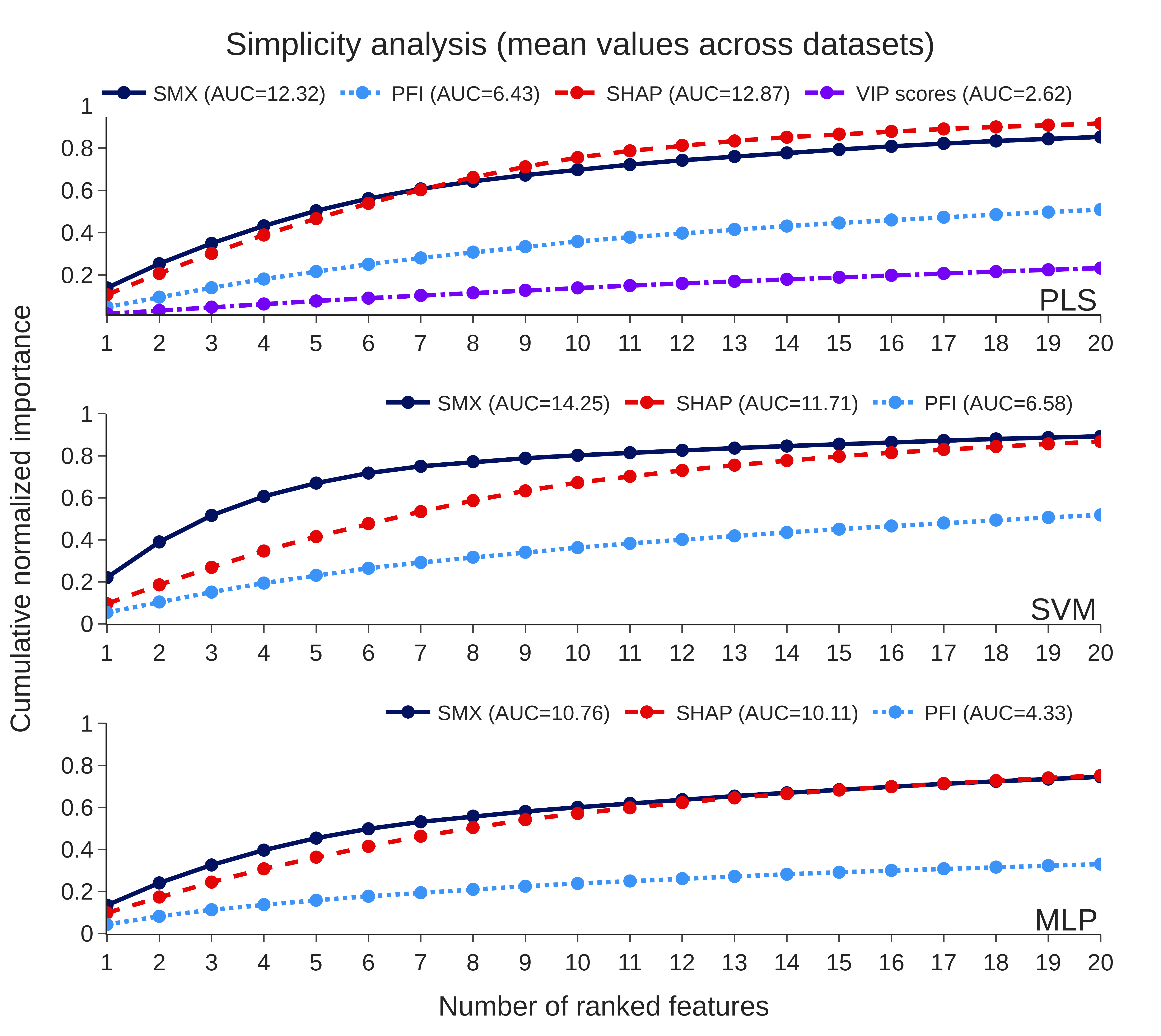}
  \caption{\footnotesize Simplicity analysis via cumulative importance proportion. The plots show the mean cumulative importance proportion across the datasets (\textit{n=}8) for each method. The x-axis indicates the number of features (variables for SHAP/PFI/VIP and predicates for SMX) sorted in descending order of importance. The legends indicate the mean area under the curve (AUC) values}
  \label{fig:simplicity}
\end{figure}

Considering simplicity, Figure~\ref{fig:simplicity} shows mean cumulative importance proportions across datasets (individual results are available in the supplementary materials). SMX and SHAP produced the highest AUC values and therefore the simplest explanations, with close curve trajectories across models. In practical terms, both methods concentrated relevance in fewer elements, whereas PFI and VIP spread importance more diffusely across larger feature sets. Wilcoxon signed-rank tests on the dataset level (Table~\ref{tab:wilcoxon_simplicity_smx}) confirm this pattern, as SMX was significantly simpler than PFI in all models ($p-values=0.008$), significantly simpler than VIP in PLS ($p-value=0.008$), and statistically non-differentiable from SHAP in all models ($p-value\geq 0.05$) within the sampled context. Overall, these evidences indicate that SMX attains high simplicity while remaining equivalent with SHAP, which is a well-established baseline.

Conversely, simplicity alone does not guarantee explanation quality, as format, organization, and structure also play a crucial role in their compositional level. Here, SMX offers a practical advantage for spectral data that is not directly reproduced by the other methods. SHAP, PFI, and VIP primarily return feature-level scores, usually requiring additional aggregation and visualization (\textit{e.g.}, bar plots, line plots, or heatmaps). SMX instead produces predicate-based outputs already linked to spectral zones, reducing the distance between algorithmic output and domain interpretation. These predicates are logical rules that define threshold conditions over spectral variables and delimit model-relevant sample subsets. As a result, SMX communicates not only \textit{where} relevance is located (zone), but also \textit{how} it contributes (condition/range) and \textit{from which} instances (most impactful subset) it emerges, adding contextual
information that raw importance magnitudes alone do not provide. Although predicates are generated in PCA space (dimensionless), they are converted back to threshold ranges in natural spectral shapes and units and overlaid on the original signal, uncovering spectral boundaries that are related to the model’s perspective. Accordingly, this mapping enables direct inspection in the same coordinate system used in laboratory interpretation, supporting chemical/physical reasoning and more actionable conclusions, fueling insights and hypothesis generation.

\begin{table}[]
  \centering
  \scriptsize
  \setlength{\tabcolsep}{6pt}
  \caption{\small Wilcoxon signed-rank results for simplicity, focusing on comparisons involving SMX across datasets (\textit{n=}8). \textbf{Median diff} corresponds to the paired median of SHAP, PFI or VIP $-$ SMX}
  \label{tab:wilcoxon_simplicity_smx}
  \begin{tabular}{llllr}
    \toprule
    \textbf{Model} & \textbf{Comparison} & \textbf{Median diff} & \textbf{p-value} \\
    \midrule
    PLS & PFI$-$SMX & -0.306 & 0.008 \\
    PLS & SHAP$-$SMX & 0.063 & 0.078 \\
    PLS & VIP$-$SMX & -0.531 & 0.008 \\
    SVM & PFI$-$SMX & -0.202 & 0.008 \\
    SVM & SHAP$-$SMX & 0.089 & 0.641 \\
    MLP & PFI$-$SMX & -0.341 & 0.008 \\
    MLP & SHAP$-$SMX & -0.022 & 0.945 \\
    \bottomrule
  \end{tabular}
\end{table}

\subsection{Case study: soil fertility using XRF spectra}
\label{sec:case_study}

After establishing the competitive performance and methodological properties of SMX, the next step is to illustrate how these features translate into practical explainability in a real spectral analysis scenario. As a showcase, this section presents a study case on the soil (XRF) dataset, more detailedly comparing explanations from SMX, SHAP, PFI, and VIP. This Brazilian dataset represents a realistic and challenging spectral classification task, where the objective is to discriminate eutrophic (more fertile) from dystrophic (less fertile) soils based on XRF signatures. The distinction is a fundamental pedological classification in Brazilian and tropical soil science that reflects differences in natural fertility, weathering history, and parent material, being based on the Base Saturation Percentage (BSP\%). BSP\% is computed from exchangeable Calcium, Magnesium, Potassium, and potential acidity (H+Al) as the Sum of Bases (SB) over the Cation Exchange Capacity (CEC) in percentage, with the conventional thresholds BSP\% $>$ 50.0\% (eutrophic) and BSP\% $\leq$ 50.0\% (dystrophic)~\cite{solos2006sistema,ribeiro2025impact}. As the reference for regular soil assessment, these soil properties are measured in laboratory through wet-chemistry procedures that, although accurate, are time-consuming, costly, and dependent on hazardous reagents. Accordingly, the development of rapid, cost-effective, and greener alternatives within this context is a key research topic in precision agriculture and environmental monitoring, where XRF-based machine learning modeling emerges as a promising solution.

As shown in Figure~\ref{fig:xrf_spectra}, the spectra carry useful Ca-derived signatures, proportional to its total content in the samples, which is particularly relevant for discriminating the two soil classes, since Ca directly relates to the BSP\%. However, such features are superimposed with overlapping contributions from other elements and background regions, making the differentiation non-trivial and explains the moderate-to-high performance levels reported in Table~\ref{tab:ml_performance}.

For this study case, we focus on PLS because all compared explainers are available (including VIP). Table~\ref{tab:pls_da_entities} reports the top-10 explanatory entities per method (full rankings are provided in the supplementary materials). A key point is that the four methods produced scores with different numerical behavior and granularity. SHAP and PFI are variable-level methods, so neighboring variables within the same spectral zone might receive close scores due to the inherent adjacency correlation structure of spectral data, which produces local collinearity among related features. In practice, this created dense rankings where discriminating adjacent positions required more decimal places. By contrast, SMX ranks predicates over aggregated zones, and VIP summarizes variable relevance through latent variables, with both tending to produce a wider score spacing in the top ranks, usually distinguishable with fewer decimals. Furthermore, variable-level methods (SHAP, PFI, and VIP) naturally generate long lists (proportional to the number of spectral variables), whereas SMX is intrinsically simpler because it ranks predicates tied to a finite set of zones and thresholds. Consequently, SMX can preserve the main explanatory signal while reducing ranking length and cognitive load, which is advantageous for human interpretability in high-dimensional spectra.

Despite these differences, the methods converge on the most relevant regions (Table~\ref{tab:pls_da_entities}). Top entities are concentrated in Ca k$\alpha$, Mn, Si, and Fe k$\alpha$ zones, with Ca k$\alpha$ occupying the first positions across methods. The main divergence is the diversification after repeated Ca selections, as PFI, SHAP, and VIP keep selecting adjacent Ca variables for longer, whereas SMX introduces cross-zone predicates earlier (\textit{e.g.}, Mn and Si). In this example, VIP is the least diverse in the top-10, concentrating on Ca k$\alpha$ and Fe k$\alpha$.

\begin{table}[]
  \centering
  \scriptsize
  \setlength{\tabcolsep}{2.2pt}
  \caption{\small Top-10 explanatory entities for the PLS model on the soil (XRF) dataset, according to each compared method}
  \label{tab:pls_da_entities}
  \resizebox{\linewidth}{!}{%
  \begin{tabular}{ccccccccc}
    \toprule
    \textbf{Rank} & \multicolumn{2}{c}{\textbf{SMX}} & \multicolumn{2}{c}{\textbf{PFI}} & \multicolumn{2}{c}{\textbf{SHAP}} & \multicolumn{2}{c}{\textbf{VIP}} \\
    \cmidrule(lr){2-3} \cmidrule(lr){4-5} \cmidrule(lr){6-7} \cmidrule(lr){8-9}
    & \textbf{Predicate} & \textbf{LRC} & \textbf{Variable (zone)} & \textbf{Score} & \textbf{Variable (zone)} & \textbf{Score} & \textbf{Variable (zone)} & \textbf{Score} \\
    \midrule
    1 & Ca k$\alpha$ $>$ -0.36 & 9.40 & 3.70 (Ca k$\alpha$) & 0.0331 & 3.70 (Ca k$\alpha$) & 0.0267 & 3.70 (Ca k$\alpha$) & 6.95 \\
    2 & Ca k$\alpha$ $>$ -1.02 & 6.75 & 3.72 (Ca k$\alpha$) & 0.0302 & 3.72 (Ca k$\alpha$) & 0.0236 & 3.68 (Ca k$\alpha$) & 6.75 \\
    3 & Mn $>$ 0.64 & 6.20 & 3.68 (Ca k$\alpha$) & 0.0296 & 3.68 (Ca k$\alpha$) & 0.0234 & 3.72 (Ca k$\alpha$) & 6.59 \\
    4 & Si $\leq$ -0.66 & 5.90 & 3.66 (Ca k$\alpha$) & 0.0240 & 3.66 (Ca k$\alpha$) & 0.0178 & 3.66 (Ca k$\alpha$) & 5.89 \\
    5 & Ca k$\alpha$ $>$ -1.73 & 5.51 & 1.74 (Si) & 0.0204 & 5.90 (Mn) & 0.0131 & 3.74 (Ca k$\alpha$) & 5.53 \\
    6 & Si $>$ 0.13 & 5.07 & 6.42 (Fe k$\alpha$) & 0.0194 & 6.42 (Fe k$\alpha$) & 0.0123 & 3.64 (Ca k$\alpha$) & 4.74 \\
    7 & Mn $>$ -0.15 & 4.17 & 5.90 (Mn) & 0.0187 & 3.64 (Ca k$\alpha$) & 0.0122 & 6.44 (Fe k$\alpha$) & 4.63 \\
    8 & Si $\leq$ -0.20 & 3.87 & 4.46 (Ti k$\alpha$) & 0.0186 & 3.74 (Ca k$\alpha$) & 0.0111 & 6.46 (Fe k$\alpha$) & 4.56 \\
    9 & Ca k$\alpha$ $\leq$ -1.02 & 3.38 & 3.64 (Ca k$\alpha$) & 0.0184 & 5.92 (Mn) & 0.0110 & 6.42 (Fe k$\alpha$) & 4.51 \\
    10 & Si $>$ -0.66 & 3.34 & 3.74 (Ca k$\alpha$) & 0.0180 & 1.74 (Si) & 0.0107 & 6.48 (Fe k$\alpha$) & 4.34 \\
    \bottomrule
  \end{tabular}%
  }
  \scriptsize{LRC: Local Reaching Centrality. In Variable (zone), the value corresponds to energy (keV) and the parenthesized label corresponds to the spectral zone.}
\end{table}

To compare methods on a common semantic unit, variable-level and predicate-level outputs were converted into ranked zone lists using the procedure described in section~\ref{sec:XAI_metrics_and_stats}. The resulting top-10 rankings are shown in Table~\ref{tab:pls_da_zones}. This conversion is intentionally lossy: when adjacent variables collapse into one zone, small feature-level order differences can become larger position shifts at zone level. Accordingly, these lists should be interpreted as high-level summaries, not exact projections of the original rankings. The results (Table~\ref{tab:pls_da_zones}) indicated strong agreements at the top. All methods share the same four dominant zones (Ca k$\alpha$, Mn, Si, and Fe k$\alpha$), differing mainly in order. Greater variability at deeper ranks is expected, because lower-importance zones tend to have closer scores and are therefore more sensitive to minor fluctuations.

\begin{table}[]
  \centering
  \scriptsize
  \setlength{\tabcolsep}{6pt}
  \caption{\small Top-10 ranked spectral zones for the PLS model on the soil (XRF) dataset after removing repeated zones (first occurrence kept)}
  \label{tab:pls_da_zones}
  \begin{tabular}{ccccc}
    \toprule
    \textbf{Rank} & \textbf{SMX} & \textbf{PFI} & \textbf{SHAP} & \textbf{VIP} \\
    \midrule
    1 & Ca k$\alpha$ & Ca k$\alpha$ & Ca k$\alpha$ & Ca k$\alpha$ \\
    2 & Mn & Si & Mn & Fe k$\alpha$ \\
    3 & Si & Fe k$\alpha$ & Fe k$\alpha$ & Mn \\
    4 & Fe k$\alpha$ & Mn & Si & Si \\
    5 & Ti k$\alpha$ & Ti k$\alpha$ & Ti k$\alpha$ & Fe k$\beta$ \\
    6 & K & Fe k$\beta$ & Fe k$\beta$ & Ca k$\beta$ \\
    7 & Al & Al & K & Ti k$\alpha$ \\
    8 & Fe k$\beta$ & Ti k$\beta$ & sum Fe & Al \\
    9 & Ca k$\beta$ & K & P & K \\
    10 & P & Ca k$\beta$ & Background 2 & sum Fe \\
    \bottomrule
  \end{tabular}
\end{table}

To quantify this top-weighted agreement, we computed pairwise RBO similarities ($\rho$=0.7; depth = 20) between the zone rankings, producing Table~\ref{tab:rbo_similarity_matrix}. All pairwise values were high (0.80--0.92), indicating convergence on the dominant explanatory signal despite differences at deeper ranks. The strongest agreement is between SMX and SHAP (0.92), while agreement with PFI (0.82) and VIP (0.80) remains substantial despite different methodological principles. Taken together, Tables~\ref{tab:pls_da_entities}, \ref{tab:pls_da_zones}, and \ref{tab:rbo_similarity_matrix} indicate that SMX is consistent with established explainers while offering a more compact, zone-centered representation.

\begin{table}[]
  \centering
  \scriptsize
  \setlength{\tabcolsep}{8pt}
  \caption{\small RBO similarity matrix comparing the ranked zone lists of the different explainability methods for the PLS model on the soil dataset (XRF). The RBO similarity ($\rho$=0.7) emphasizes the top ranks in the comparison so that higher values indicate greater similarity between the ranked lists, with 1.0 representing identical rankings}
  \label{tab:rbo_similarity_matrix}
  \begin{tabular}{lcccc}
    \toprule
    & \textbf{SMX} & \textbf{PFI} & \textbf{SHAP} & \textbf{VIP} \\
    \midrule
    \textbf{SMX} & 1.00 & - & - & - \\
    \textbf{PFI} & 0.82 & 1.00 & - & - \\
    \textbf{SHAP} & 0.92 & 0.82 & 1.00 & - \\
    \textbf{VIP} & 0.80 & 0.80 & 0.85 & 1.00 \\
    \bottomrule
  \end{tabular}
\end{table}

\begin{figure*}[]
  \centering
  \includegraphics[width=1.0\textwidth]{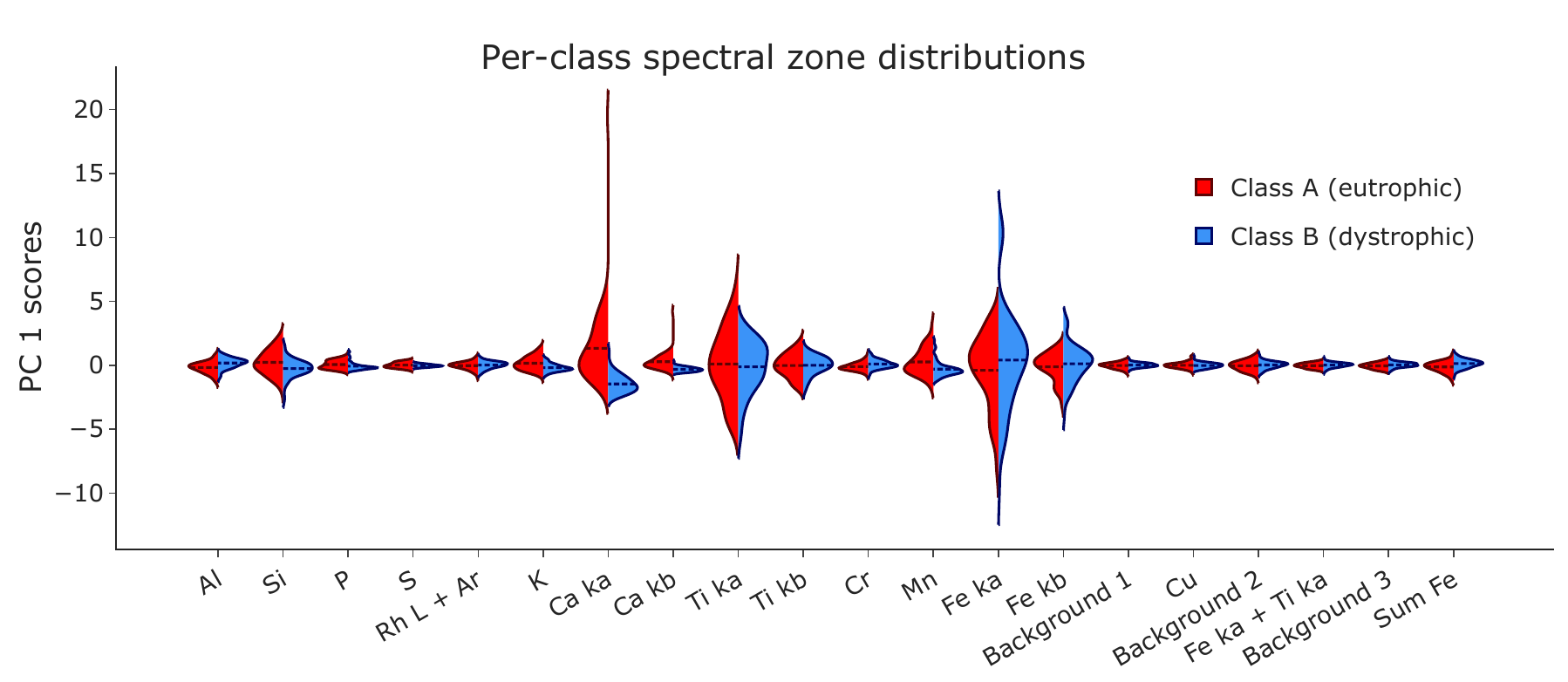}
  \caption{\small Violin plots showing the distribution of PC 1 scores for the spectral zones of the soil (XRF) dataset}
  \label{fig:scores_distribution}
\end{figure*}

Having quantitatively and qualitatively compared SMX and baselines, we now examine the practical insights uniquely enabled by SMX. Comparing per-class score distributions (Figure~\ref{fig:scores_distribution}) with the ranking in Table~\ref{tab:pls_da_zones} shows that top-ranked zones (\textit{e.g.}, Ca k$\alpha$, Mn, Si, and Fe k$\alpha$) are those with clearer class separation. This is consistent with the faithfulness-alignment mechanism of SMX, which prioritizes zones where perturbations produce stronger shifts in model output.

From a soil-fertility standpoint, the SMX-derived ranking indicate that the model learned from agronomically coherent features, since the first ranked spectral zones arise from elements intrinsically connected to BSP\%, which is the reference indicator for the eutrophic/dystrophic classification. Calcium confirmed its importance as the most relevant element, reflecting its central relation to the BSP\% through SB and CEC~\cite{solos2006sistema}. Soil acidity, which is inversely related to fertility, linked to CEC and therefore co-varies with BSP\%, regulates the solubility and availability of several elements, including Fe and specially Mn, frequently leading to their increased availability and toxicity at lower pH levels~\cite{hodges2010soil}. Silicon, in turn, acts as a structural contributor to soil mineralogy and texture (Clay vs Sand), indirectly influencing fertility and thereby also co-varying with BSP\%. Therefore, the SMX ranking is not only statistically grounded but also consistent with soil science principles.

\begin{figure}[]
  \centering
  \includegraphics[width=1.0\linewidth]{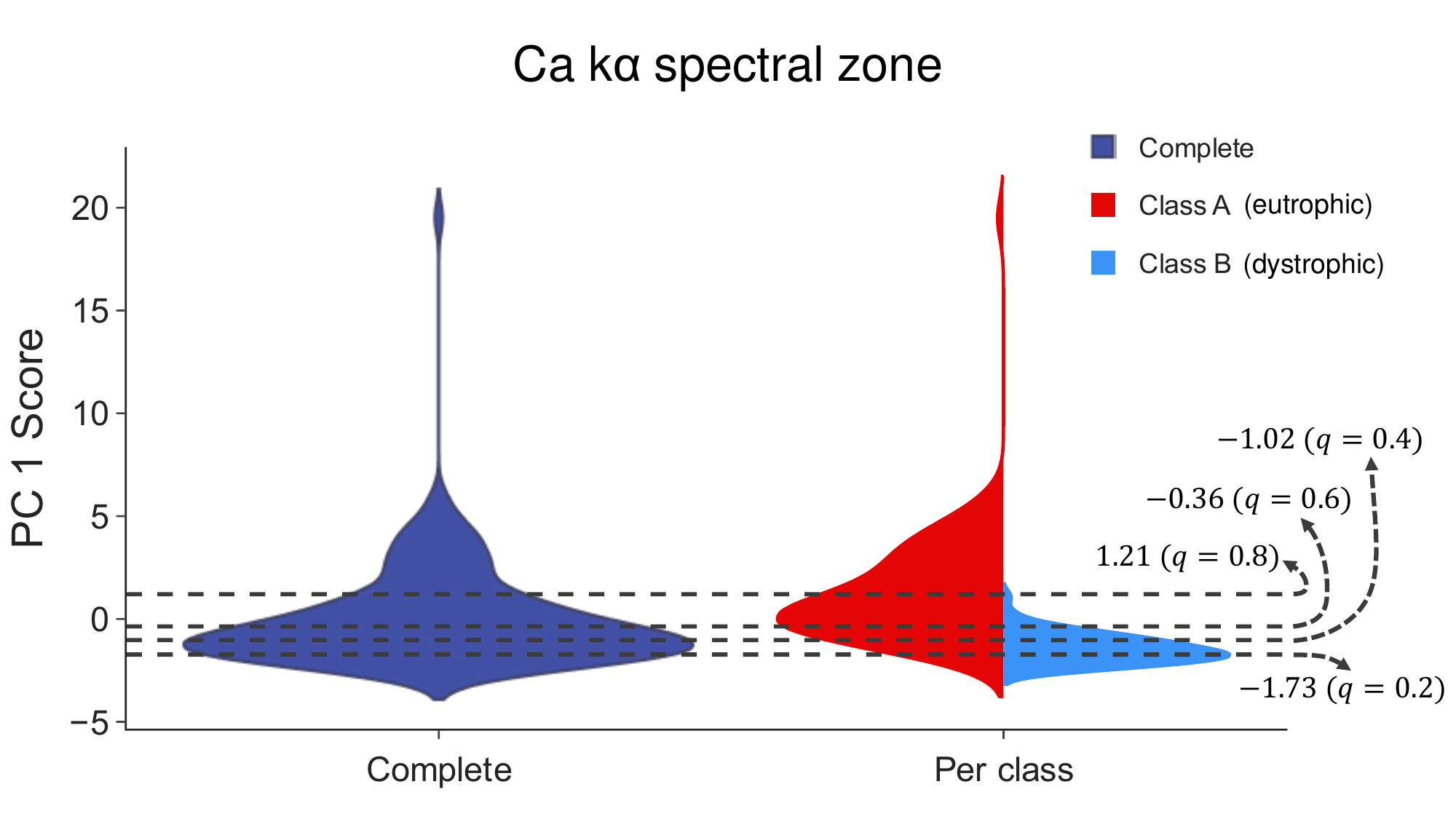}
  \caption{\small Violin plots showing the distribution of PCA scores for the Ca k$\alpha$ zone. The horizontal dashed line indicates the threshold defined by the employed quantiles (\textit{q=}[0.2, 0.4, 0.6, 0.8]) for drawing the predicates, computed on the complete training set}
  \label{fig:scores_distribution_ca}
\end{figure}

On the other hand, predicates represent operational statements about model-relevant subpopulations within each zone. This is illustrated for Ca k$\alpha$ in Figure~\ref{fig:scores_distribution_ca}, where the violin plots depict the per-zone PCA score distributions, with the dashed horizontal lines marking the quantile thresholds (\textit{q}=[0.2, 0.4, 0.6, 0.8]) computed on the training set. These class distributions explain why this zone is the most relevant for the model's predictions: eutrophic soils occupy broader and generally higher score ranges, while dystrophic soils concentrate at lower values. Moreover, each quantile vertically defines a different partition, and therefore a different explanatory lens. Higher quantiles (\textit{e.g.}, $q=0.8$) produced purer subsets, whereas lower quantiles (\textit{e.g.}, $q=0.2$) produced larger, more heterogeneous subsets. Here, a distinction is important: SMX does not rank predicates by class purity, as highly pure subsets not necessarily correspond to the most influential ones from the model’s perspective (unless the models' learning nature is grounded on identifying pure data partitions, \textit{e.g}., decision trees, random forests, and boost-based variants). Instead, SMX seeks the thresholds that most shift the model outputs when the corresponding zones are perturbed according to the delimited subsets, producing a causality-oriented ranking that prioritizes predicates based on their actual impact on the model’s behavior. For instance, the predicate induced by \textit{q=0.6} outranks \textit{q=0.8} (Table~\ref{tab:pls_da_entities}) because it causes a stronger output shift.

\begin{figure*}[] \centering \includegraphics[width=1.0\textwidth]{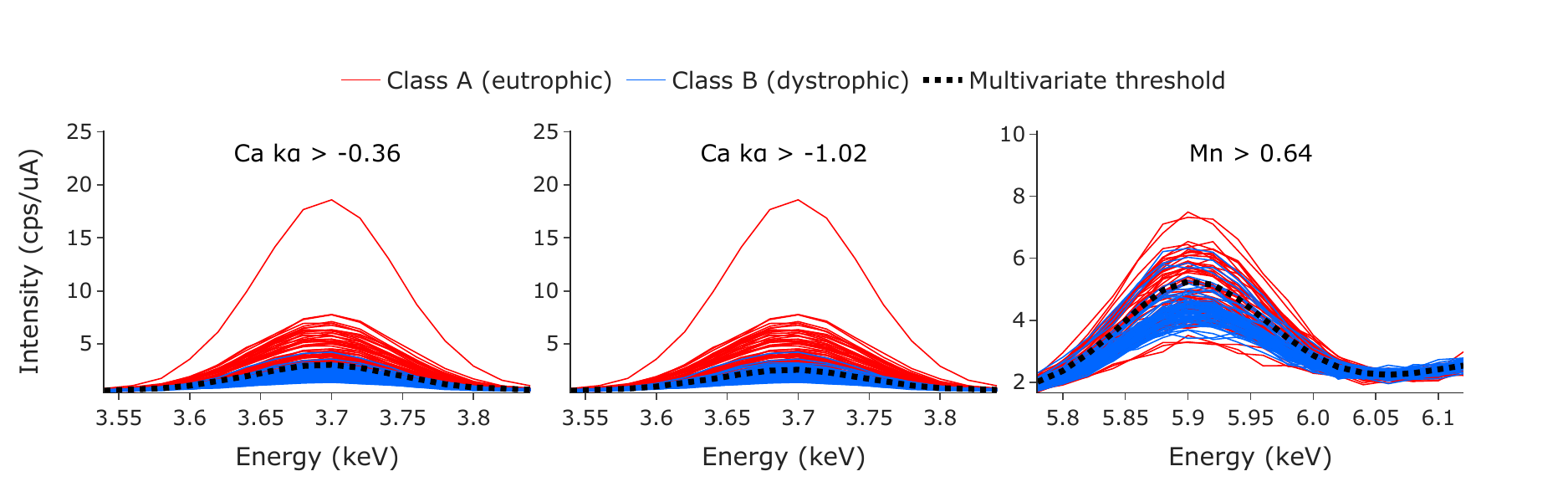} \caption{\small Measured XRF spectra of eutrophic (Class~A, red) and dystrophic (Class~B, blue) soils overlaid with the threshold spectra $\boldsymbol{\tau}^{\mathrm{spectrum}}$ (dotted curves, Eq.~\ref{eq:threshold_spectrum}) of the three top-ranked SMX predicates for the PLS model. Each dotted curve represents the spectral profile of a sample lying exactly on the predicate boundary: spectra above it satisfy the $>$ condition (predominantly eutrophic) and spectra below satisfy the $\leq$ condition (predominantly dystrophic). Left: Ca~k$\alpha > -0.36$ (LRC~$= 9.40$), Center: Ca~k$\alpha > -1.02$ (LRC~$= 6.75$), and Right: Mn~$> 0.64$ (LRC~$= 6.20$)} \label{fig:top3_predicate} \end{figure*}

Beyond identifying relevant sample subsets, the predicate's threshold values themselves live in PCA space and are not directly interpretable in natural spectral units. SMX addresses this by back-projecting them to the original domain as multivariate, spectral thresholds. Figure~\ref{fig:top3_predicate} shows this reconstruction overlaid on measured spectra for the three top-ranked predicates, translating abstract conditions into an easily readable boundaries. In practical terms, Ca k$\alpha$ $>$ -0.36, Ca k$\alpha$ $>$ -1.02, and Mn $>$ 0.64 correspond to samples above the boundaries (or below if the predicates involved $\leq$), predominantly eutrophic and associated with higher Ca and Mn signal intensities. This step is central to SMX's contribution, since it preserves faithfulness to model behavior (via perturbation-based relevance) while improving the explanations' domain alignment and composition (via interpretation in the specialist's coordinate system).

\subsection{Threshold spectrum reconstruction: analytical 
interpretation}
\label{sec:threshold_interpretation}

This section fully discusses the physical interpretation of the predicate boundaries and extends their utility from model interpretation to concrete analytical workflows. 


\subsubsection{Physical meaning of the threshold spectrum}

The threshold spectrum $\boldsymbol{\tau}^{\mathrm{spectrum}}$ corresponding to a predicate $P_{m,k}$ is defined in Eq.~\ref{eq:threshold_spectrum} as the spectral profile lying exactly on the decision boundary of that predicate within zone~$Z_m$. Concretely, it is the multivariate spectral signal 
that, when projected onto the first principal component of zone~$Z_m$, yields a score exactly equal to the quantile threshold~$\tau_{m,k}$. Samples whose zone score exceeds $\tau_{m,k}$ (predicate $P^{>}_{m,k}$) produce spectra with higher intensity in the directions captured by the loading vector $\mathbf{w}^{(m)}_1$ relative to the threshold profile, while samples below $\tau_{m,k}$ (predicate $P^{\leq}_{m,k}$) produce spectra with comparatively lower intensity in those directions. The threshold spectrum therefore represents a physically meaningful boundary: it is the spectral signature of a hypothetical sample sitting exactly at the predicate frontier. Therefore, as more important the predicate is, more relevant the spectral features of its threshold spectrum are for the model's decision process.

\subsubsection{Practical utility for analytical workflows}

Beyond model interpretation, threshold spectra enable concrete applications in analytical practice that purely numerical importance scores cannot support. Some examples include:

\textit{Sample screening and class assignment.} A practitioner can overlay a newly acquired spectrum on the threshold profiles of the top-ranked predicates and assess, by direct visual inspection, whether the sample lies above or below each boundary. For well-separated classes such as eutrophic and dystrophic soils in the Ca~k$\alpha$ zone, this visual comparison may suffice to support a preliminary classification decision without invoking the full model, reducing the computational overhead for routine screening.

\textit{Prioritisation of confirmatory analysis.} Samples whose spectra lie close to a predicate threshold represent borderline cases where the opposite class may be close. Therefore, spectra that nearly coincide with the dotted boundary in one or more top-ranked zones may be prioritized for confirmatory wet-chemistry analysis, focusing laboratory resources on pertinent cases rather than on samples that are clearly classified.

\textit{Instrument monitoring and calibration drift detection.} If the threshold spectrum derived from a calibration dataset is stored as a reference profile, systematic shifts of routine spectra relative to this reference in the high-importance zones may signal calibration drift, matrix effects, or changes in instrument response before any degradation in predictive performance becomes evident. This enables proactive quality control based on spectral boundaries that are directly tied to the model's decision logic rather than to global spectral statistics.

\textit{Hypothesis generation for domain research.} When threshold spectra recur consistently across different quantile levels or datasets, their spectral position and intensity provide a structured, evidence-based starting point for hypotheses about the physicochemical mechanisms underlying class separation. In the soil fertility example, the convergence of Ca~k$\alpha$ and Mn threshold profiles across multiple predicates is consistent with the joint role of Ca and Mn in determining BSP\% and soil pH, respectively, and could motivate targeted geochemical investigations into the co-variation of these elements across tropical soil types.


\subsection{SMX hyperparameters sensitivity}
\label{subsec:ablation_study}

To further clarify the sensitivity of SMX to its key hyperparameters, an ablation study was conducted on the synthetic dataset. As shown in Table~\ref{tab:ml_performance}, the synthetic data-based models achieved the highest performance levels (Accuracy, Sensitivity, and Specificity $=1.0$). Accordingly, the SMX-extracted explanations following the baseline settings (Table~\ref{tab:ablation_predicates}) delivered a list of predicates hierarchically consistent with the data generation logic, where the top-ranked features reflect the granular conditions defining the most influential spectral zones. For instance, predicates corresponding to the Feature 1 (the zone carrying the strongest discriminative signals) occupied the top positions and exhibited higher LRC scores, while predicates firstly related to Feature 2 and then to Feature 3 (zones with the second and third strongest signals, respectively) appeared in subsequent ranks, followed by background-related variables. 

\begin{table}[]
  \centering
  \scriptsize
  \setlength{\tabcolsep}{2.2pt}
  \caption{\small Top-20 ranked predicates for the synthetic dataset in the PLS, SVM, and MLP models. The predicates are ordered by their Local Reaching Centrality (LRC) values, where the rounding was atypically set to 6 due to the proximity of the scores of the related to Feature 3 and background variables}
  \label{tab:ablation_predicates}
  \resizebox{\linewidth}{!}{%
  \begin{tabular}{ccccccc}
    \toprule
    \textbf{Rank} & \textbf{PLS} & \textbf{LRC} & \textbf{SVM} & \textbf{LRC} & \textbf{MLP} & \textbf{LRC} \\
    \midrule
    1 & Feat 1 $>$ 3.27 & 8.553126 & Feat 1 $>$ 3.27 & 13.046961 & Feat 1 $>$ 3.27 & 10.887024 \\
    2 & Feat 1 $>$ -4.02 & 6.221655 & Feat 1 $>$ -4.02 & 9.086454 & Feat 1 $>$ -4.02 & 7.678767 \\
    3 & Feat 1 $>$ -4.13 & 4.186363 & Feat 1 $>$ -4.13 & 5.495456 & Feat 1 $>$ -4.13 & 4.800323 \\
    4 & Feat 1 $\leq$ 4.66 & 2.452428 & Feat 1 $\leq$ 4.66 & 2.683337 & Feat 1 $\leq$ 4.66 & 2.529788 \\
    5 & Feat 1 $\leq$ 3.27 & 2.029708 & Feat 1 $\leq$ 3.27 & 0.891648 & Feat 1 $\leq$ 3.27 & 1.083680 \\
    6 & Feat 2 $>$ 1.07 & 1.768245 & Feat 2 $>$ 1.07 & 0.143849 & Feat 2 $>$ 1.07 & 0.455096 \\
    7 & Feat 2 $\leq$ -1.36 & 1.550746 & Feat 2 $>$ -1.36 & 0.117657 & Feat 2 $>$ -1.36 & 0.343913 \\
    8 & Feat 2 $>$ -1.36 & 1.263268 & Feat 2 $>$ -2.24 & 0.093897 & Feat 1 $\leq$ -4.02 & 0.328563 \\
    9 & Feat 2 $>$ -2.24 & 0.985201 & Feat 1 $\leq$ -4.02 & 0.086908 & Feat 2 $>$ -2.24 & 0.312550 \\
    10 & Feat 2 $\leq$ 2.23 & 0.688141 & Feat 2 $\leq$ 2.23 & 0.082857 & Feat 2 $\leq$ -1.36 & 0.280756 \\
    11 & Feat 2 $\leq$ 1.07 & 0.498153 & Feat 3 $>$ 0.37 & 0.079914 & Feat 2 $\leq$ 2.23 & 0.208652 \\
    12 & Feat 1 $\leq$ -4.02 & 0.212261 & Feat 2 $\leq$ -1.36 & 0.073188 & Feat 2 $\leq$ 1.07 & 0.131328 \\
    13 & Feat 3 $>$ 0.37 & 0.015712 & Feat 2 $\leq$ 1.07 & 0.061942 & Feat 3 $\leq$ -0.32 & 0.069232 \\
    14 & Feat 3 $\leq$ -0.32 & 0.013940 & Feat 3 $>$ -0.32 & 0.049132 & Feat 3 $>$ 0.37 & 0.060472 \\
    15 & Feat 3 $\leq$ 0.37 & 0.008962 & Feat 3 $>$ -1.13 & 0.039224 & Feat 3 $\leq$ 0.37 & 0.050569 \\
    16 & Feat 3 $>$ -0.32 & 0.008270 & Feat 3 $\leq$ -0.32 & 0.029456 & Feat 3 $\leq$ 1.11 & 0.043085 \\
    17 & Feat 3 $\leq$ 1.11 & 0.007539 & Feat 3 $\leq$ 1.11 & 0.018940 & Feat 3 $>$ -0.32 & 0.031364 \\
    18 & Feat 3 $>$ -1.13 & 0.005767 & Feat 3 $\leq$ 0.37 & 0.006249 & Feat 3 $>$ -1.13 & 0.015690 \\
    19 & Back 2 $\leq$ -0.01 & 0.000151 & Back 4 $\leq$ -0.03 & 0.000395 & Back 2 $\leq$ -0.01 & 0.002258 \\
    20 & Back 2 $>$ 0.02 & 0.000121 & Back 4 $>$ 0.04 & 0.000336 & Back 2 $\leq$ 0.02 & 0.001472 \\
    \bottomrule
  \end{tabular}
  }
  \scriptsize{LRC: Local Reaching Centrality, Feat: Features, Back: Background}  
\end{table}

\begin{figure*}[]
  \centering
  \includegraphics[width=0.9\linewidth]{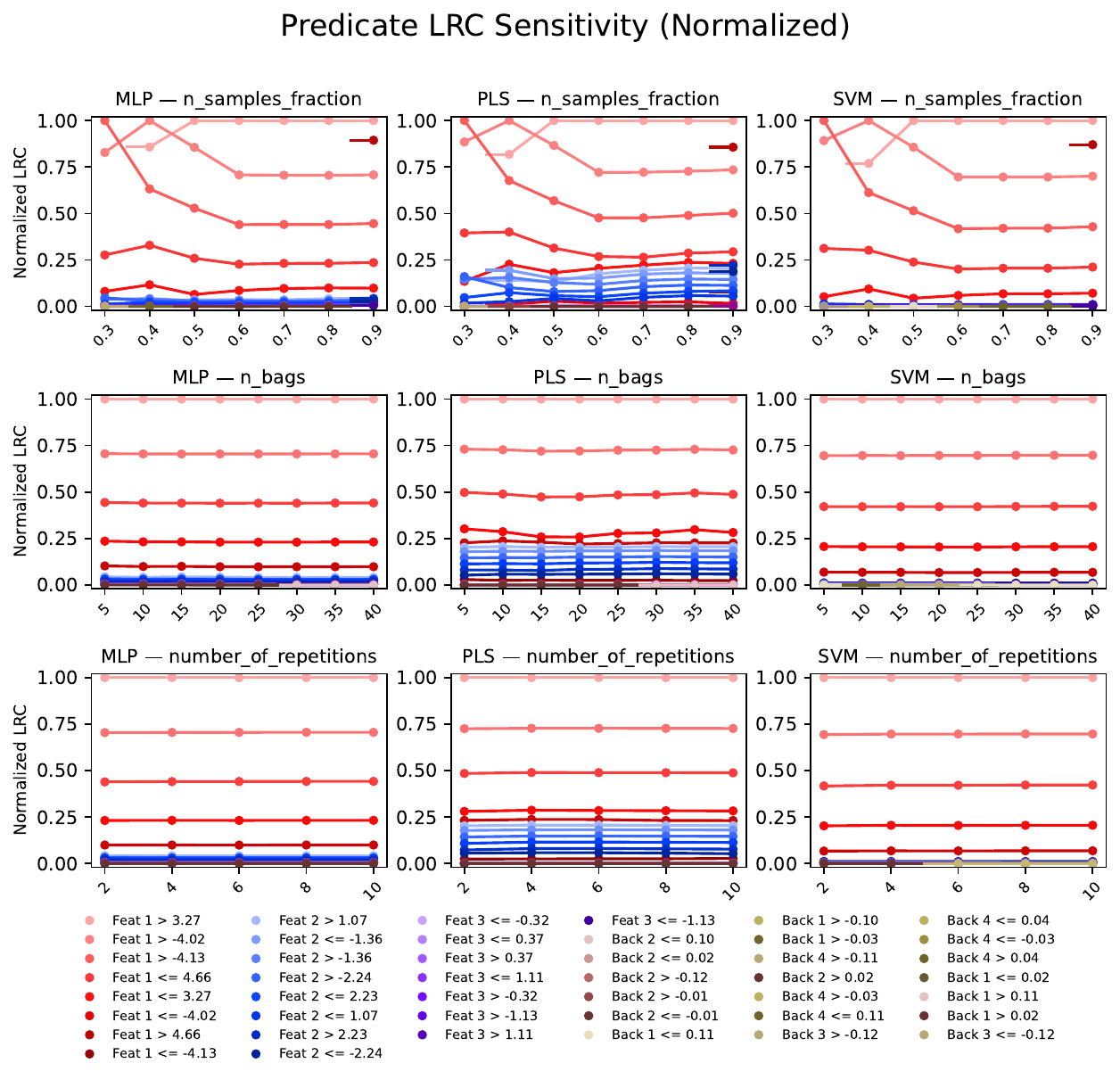}
  \caption{\small Ablation study on the number of bags hyperparameter employing the synthetic dataset. LRC: Local Reaching Centrality, Feat: Features, Back: Background}
  \label{fig:rank_sensitivity_hyperparameters}
\end{figure*}

Based on these baseline models, the SMX hyperparameters were varied one at a time to evaluate their influence on the extracted explanations, following the procedure described in detail in section~\ref{sec:ablation}. The results for the number of bags, repetitions and samples per bag are reported in Figure \ref{fig:rank_sensitivity_hyperparameters}, where the predicates related to Feature 1 and Feature 2 consistently occupied the top positions, while those centered on Feature 3 and background zones occupied the bottom ones and exhibited more pronounced rank fluctuations across the different hyperparameter settings, as expected given their close-to-zero, tightly-spaced LRC levels and therefore higher sensitivity to the hyperparameter. Shifts in the first positions (predicates based on Features 1 and 2) were observed when varying the number of samples per bag, mainly when small-to-medium fractions of the training set were employed (0.3--0.6 \%). This is consistent with the fact that smaller bags yield sample subsets with higher variance, which can lead to more variable perturbation effects and therefore more fluctuations in the LRC-based rankings. Again, the most affected predicates were those related to Feature 3 and background zones. Despite that, even in these cases, the data generation hierarchical consistency of the top-ranked predicates was preserved.

\begin{figure*}[]
  \centering
  \includegraphics[width=0.9\linewidth]{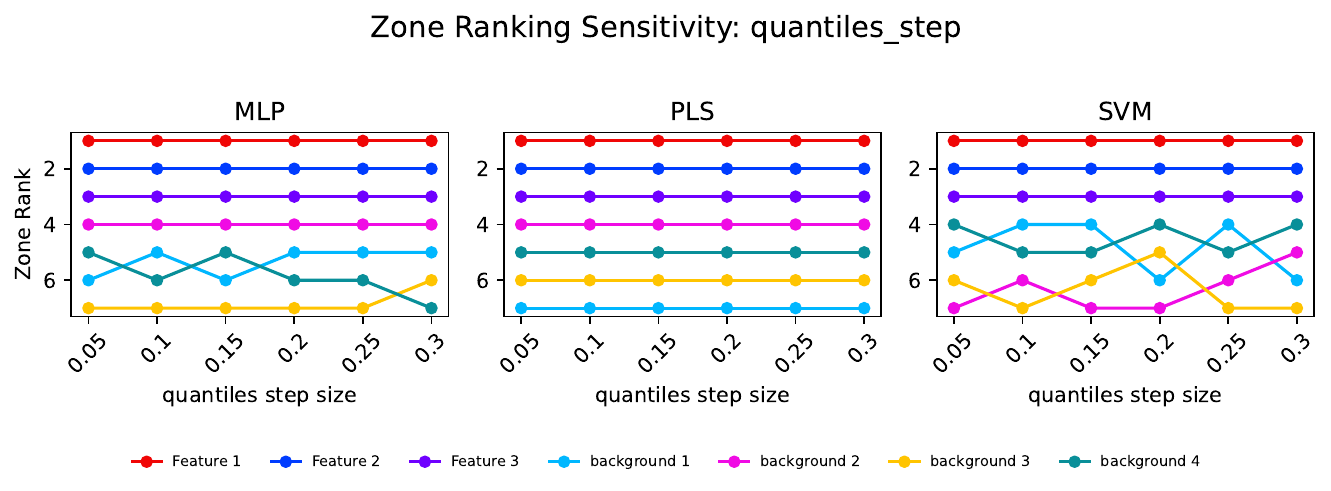}
  \caption{\small Ablation study on the quantiles hyperparameter employing the synthetic dataset}
  \label{fig:rank_sensitivity_quantiles_step}
\end{figure*}

Regarding the quantiles, the analysis is different, since they define the predicates themselves through the thresholds, so their variation produces distinct sets of explanations. Aggregating them into their corresponding zones and comparing the resulted rankings (Figure~\ref{fig:rank_sensitivity_quantiles_step}) showed that the referenced zones (Feature 1, Feature 2, and Feature 3) consistently occupied the top positions according to the expected order throughout the different quantile settings, with rank fluctuations centered on the background zones, which is consistent with the previous analyses. This indicates that the most relevant features were consistently ranked regardless of the specific thresholds employed to define the predicates, supporting the robustness of SMX to this hyperparameter. 

According to these findings, we recommend adopting a moderate number of bags (\textit{e.g.}, 5--10) and repetitions (\textit{e.g.}, 4--6), as this configuration offers a practical balance between stability and computational cost in SMX experiments. The number of samples per bag, conversely, should be set cautiously to limit excessive variance in perturbation effects but yet allow for sampling variability that supports robust predicate relevance estimation. This is especially important when the modeling task is expected to involve equally or closely important spectral zones, as the resulting predicates will have close LRC scores and therefore be more sensitive to sampling fluctuations. Therefore, we recommend using per-bag sample numbers directly linked to the training set size (\textit{e.g.}, 70--90 \% of the samples).

In contrast, the choice of quantile thresholds should be guided by the level of precision required in each application. More closely spaced quantile thresholds along the score distribution produce finer-grained sets of explanations by splitting the data into narrower and more specific subsets, which supports a more detailed connection to the model behavior and purer class-related boundaries. However, this granularity lead to spending more resources in computing and analyzing a larger number of predicates, as their maximum number grows with the number of quantiles as $n_{predicates}=2 \times n_{zones} \times n_{quantiles}$ (see section \ref{sec:predicates}). More widely spaced thresholds, in turn, produce a more concise set of explanations by creating broader subsets that are less computationally demanding but might provide less discriminative multivariate threshold curves when back-projected to the original domain. Therefore, we recommend a balance between granularity and conciseness (\textit{e.g.}, \textit{q=}[0.2, 0.4, 0.6, 0.8]) when selecting quantile values, considering especially how overlapped the studied data classes are throughout the spectral zones. 

\section{Limitations and Future Directions}
\label{sec:limitations}

While SMX offers a novel and spectral-grounded approach to explain models, some limitations and opportunities for improvement deserve attention.

\textit{Task scope:}
The current implementation is restricted to binary classification. Extension to multi-class settings is conceptually direct: the perturbation impact metric can be computed per-class using one-vs-rest probability shifts, and the graph terminal nodes can be expanded to accommodate $C > 2$ class labels. Extension to regression is equally tractable at the perturbation scoring stage, where the mean absolute error between original and perturbed predictions replaces the probability shift metric, and the predicate formulation and threshold spectrum reconstruction stages require no modification. The primary methodological challenge for regression lies in the graph construction, where class-terminal nodes must be replaced by, for example, discretized output range intervals (\textit{e.g.}, defined by quantiles of the response distribution). This adaptation would make SMX directly applicable to quantitative spectroscopic tasks such as elemental concentration determination, moisture content prediction, and physical or chemical property estimation, which represent a large share of analytical chemistry applications. Experimental validation on these tasks are direct continuations of the present study.

\textit{Explanation scope:}
SMX currently delivers global explanations summarizing model behavior across all training instances. Future work could exploit the predicates' subset-awareness, each predicate already partitions samples into condition-defined subgroups, to formalize local explanation routines that characterize model behavior for individual samples or specific subpopulations, complementing the global perspective.

\textit{Zone delimitation:}
Zone boundaries are expert-defined in this first proposal, which reduces portability in domains where prior spectroscopic knowledge is limited or where characteristic signals are not well established. A promising direction is to employ data-driven zone definitions to replace or complement expert input, improving applicability in less-characterized matrices. This could include automatic segmentation via peak detection algorithms, clustering-based approaches grouping adjacent variables with similar behavior, or hybrid strategies merging expert knowledge with data-driven insights to refine zone boundaries iteratively based on model-relevant patterns.

\textit{Predicate refinement:}
Current predicate formulation relies on predefined quantiles. Although this is a straightforward approach, algorithmic strategies for seeking optimal per-zone thresholds (\textit{e.g.}, linear or binary searches) that target subsets of samples with maximum perturbation scores could enhance the relevance of the extracted predicates.

\textit{Computational cost:}
Predicate relevance estimation through repeated perturbations across bags and seeds scales linearly with the number of bags, predicates, and repetitions, and is strongly influenced by the inference cost of the explained model. For large datasets or fine quantile grids this can become demanding. Parallelization of the bag-wise perturbation loop, adaptive stopping criteria based on ranking convergence, and predicate pruning strategies that discard low-support predicates early are promising directions for improving scalability without sacrificing explanation quality.

\textit{External validation:}
The evaluation was conducted exclusively on synthetic, XRF and GRS data. Broader validation across other spectral modalities, such as vis-NIR, mid-infrared, Raman, and LIBS, is needed to consolidate the generalizability of SMX, particularly given that spectral correlation structures, peak widths, and signal-to-noise characteristics differ substantially across the spectroscopic techniques.

\section{Conclusion}

Spectral-based machine learning models are increasingly deployed in analytical chemistry workflows where predictive accuracy alone is insufficient. Predictions should be traceable to physically meaningful spectral regions to support validation, quality assurance, and scientific interpretation. This study introduced the Spectral Model eXplainer (SMX), a post-hoc, global, and model-agnostic explainability framework that addresses this need by operating natively on expert-defined spectral zones rather than on isolated variables. Through quantile-based predicate formulation, stochastic bagging, perturbation scoring, and graph-based centrality analysis, SMX produces ranked zone explanations that are compact, reproducible, and expressed in the natural units of the instrument via threshold spectrum reconstruction.

Evaluated across eight real datasets including XRF and GRS spectra and three classifiers of increasing complexity (PLS, SVM, and MLP), SMX demonstrated faithfulness equivalence with baselines (SHAP, PFI, and VIP) in most settings, competitive-to-superior domain
alignment and stability, and simpler outputs than PFI and VIP while remaining comparable to SHAP (comparisons via Wilcoxon signed-rank tests).

The soil fertility case study demonstrated that SMX recovers the same dominant zones identified by established methods (Ca~k$\alpha$, Mn, Si, and Fe~k$\alpha$) while additionally providing threshold spectra that translate predicate boundaries into spectral profiles directly overlaid on instrument readings. Practical uses of these profiles may include enabling sample screening, prioritizing confirmatory wet-chemistry analysis, and detecting instrument drift within a single interpretive output, capabilities that purely numerical importance scores cannot provide.

Among SMX's contributions, the threshold spectrum reconstruction is the most analytically distinctive. By exploiting the linear invertibility of PCA, it transforms an abstract logical rule defined in score space into a measurable spectral boundary expressed in the coordinate system familiar to laboratory practitioners, bridging statistical model behavior and domain expertise in a form directly actionable for analytical decision-making.

The ablation study clarified how SMX hyperparameters affect explanation stability. Variations in the number of bags and repetitions induced only minor rank fluctuations, whereas smaller samples per bag increased variance in perturbation effects and ranking instability. Quantile settings mainly controlled explanation granularity rather than the identity of dominant zones at the aggregated level, supporting the robustness of the main explanatory signal.

Overall, SMX offers a principled, instrument-native tool for interrogating spectral binary classifiers, representing a promising step toward spectral-native explainability and opening new avenues for integrating XAI insights into practical workflows, decision support, and scientific discovery across spectroscopic applications.

\section{Acknowledgments}

The authors acknowledge the support of CNPq, Brazil (grant number 306309/2023-8, project number: 404214/2021-5) and INCT-FNA, Brazil (408419/2024-5).

\section{Declaration of generative AI and AI-assisted technologies in the writing process}

During the preparation of this study the authors used ChatGPT-4o to improve the readability and language of the manuscript. After using this tool/service, the authors reviewed and edited the content as needed and take full responsibility for the content of this publication.

\bibliographystyle{unsrtnat}
\bibliography{references}

\appendix
\section{Appendix}
\label{sec:appendix}

\subsection{Pseudocode and computational complexity}
\label{app:pseudocode}
This section provides the complete SMX implementation through the summarized Algorithms~\ref{alg:smx_phase1}--\ref{alg:smx_phase5}. The algorithm ~\ref{alg:smx_phase1} encodes the zones decomposition and aggregation by PCA, the algorithm~\ref{alg:smx_phase2} encodes the predicates formulation, the algorithm~\ref{alg:smx_phase3} encodes the bagging and perturbation scoring, and the algorithm~\ref{alg:smx_phase5} encodes the graph generation and LRC analysis.

\begin{breakablealgorithm}
\caption{SMX: Spectral-zone decomposition and aggregation}
\label{alg:smx_phase1}
\begin{algorithmic}[1]
\Require $\mathbf{X} \in \mathbb{R}^{n \times p}$, $\{Z_m\}_{m=1}^{M}$ (spectral zones with variable boundaries)
\Ensure $\mathbf{T} \in \mathbb{R}^{n\times M}$ (zone score matrix), $\{\mathbf{w}_1^{(m)}\}_{m=1}^{M}$ (zone loadings), $\{\mathrm{VE}^{(m)}\}_{m=1}^{M}$ (explained variance ratios)

\For{$m = 1$ to $M$}
    \State Extract sub-matrix $\mathbf{X}_{Z_m} \in \mathbb{R}^{n \times d_m}$ from $\mathbf{X}$
    \State Center the data: $\tilde{\mathbf{X}}_{Z_m} \leftarrow \mathbf{X}_{Z_m} - \mathbf{1}_n \bar{\mathbf{x}}_{Z_m}^{\top}$
    \State Fit PCA with one component on $\tilde{\mathbf{X}}_{Z_m}$
    \State Obtain loading vector $\mathbf{w}_1^{(m)}$
    \State Compute zone scores $t_i^{(m)}$ for all $i=1,\dots,n$
    \State Compute explained variance ratio $\mathrm{VE}^{(m)}$
\EndFor
\State Form score matrix $\mathbf{T} \leftarrow [t_i^{(m)}]_{n\times M}$
\State \Return $\mathbf{T}, \{\mathbf{w}_1^{(m)}\}_{m=1}^{M}, \{\mathrm{VE}^{(m)}\}_{m=1}^{M}$
\end{algorithmic}
\end{breakablealgorithm}
\begin{breakablealgorithm}
\caption{SMX: Predicate formulation}
\label{alg:smx_phase2}
\begin{algorithmic}[1]
\Require $\mathbf{T} \in \mathbb{R}^{n\times M}$ (zone score matrix), $\mathcal{Q} = \{q_1,\ldots,q_K\}$ (quantile levels)
\Ensure $\mathcal{P}$ (set of unique predicates), $\mathbf{I} \in \{0,1\}^{n\times N_P'}$ (indicator matrix)

\State $\mathcal{P} \leftarrow \emptyset$
\For{$m = 1$ to $M$}
    \For{$k = 1$ to $K$}
        \State Compute threshold $\tau_{m,k} = Q_{q_k}(t_1^{(m)},\ldots,t_n^{(m)})$
        \State Define $P_{m,k}^{\le}: t_i^{(m)} \le \tau_{m,k}$
        \State Define $P_{m,k}^{>}: t_i^{(m)} > \tau_{m,k}$
        \State Add $P_{m,k}^{\le}$ and $P_{m,k}^{>}$ to $\mathcal{P}$ if not duplicates
    \EndFor
\EndFor
\State Let $N_P' \leftarrow |\mathcal{P}|$
\State Build indicator matrix $\mathbf{I} \in \{0,1\}^{n\times N_P'}$ such that $I_{i,j}=1$ iff sample $i$ satisfies predicate $P_j \in \mathcal{P}$
\State \Return $\mathcal{P}, \mathbf{I}$
\end{algorithmic}
\end{breakablealgorithm}

\begin{breakablealgorithm}
\caption{SMX: Bagging and Scoring}
\label{alg:smx_phase3}
\begin{algorithmic}[1]
\Require $\mathbf{X} \in \mathbb{R}^{n \times p}$, $f$ (trained model), $\mathcal{P}$ (predicates), $\mathbf{I}$ (indicator matrix), $\{Z_m\}_{m=1}^{M}$ (spectral zones), $B$ (number of bags), $n_b$ (subsample size), $n_{\min}$ (minimum support threshold, 20 \% of the training set size), seed $r$ (varying according to the number of repetitions)
\Ensure $\{ \mathcal{L}_b \}_{b=1}^{B}$ (bag-wise ranked predicate lists), $\{\operatorname{Imp}_b(P_j)\}$ (normalized predicate impacts)

\For{$b = 1$ to $B$}
    \State Draw subsample $\mathcal{S}_b \subset \{1,\ldots,n\}$ of size $n_b$ using seed $r$
    \State Initialize valid predicate set $\mathcal{V}_b \leftarrow \emptyset$
    \For{each predicate $P_j \in \mathcal{P}$ associated with zone $Z_{m_j}$ of size $d_{m_j}$}
        \State $\mathcal{S}_b^{(j)} \leftarrow \{ i \in \mathcal{S}_b : I_{i,j}=1 \}$
        \If{$|\mathcal{S}_b^{(j)}| < n_{\min}$}
            \State \textbf{continue}
        \EndIf
        \State Perturb zone $Z_{m_j}$ for samples in $\mathcal{S}_b^{(j)}$
        \State Query $f$ on original and perturbed samples in $\mathcal{S}_b^{(j)}$
        \State Compute normalized impact
        \[
            \operatorname{Imp}_b(P_j) \leftarrow \operatorname{Imp}(P_j)/d_{m_j}
        \]
        \State Add $P_j$ to $\mathcal{V}_b$
    \EndFor
    \State Sort predicates in $\mathcal{V}_b$ by $\operatorname{Imp}_b(P_j)$ in descending order
    \State Store the resulting ranking as $\mathcal{L}_b$
\EndFor
\State \Return $\{ \mathcal{L}_b \}_{b=1}^{B}, \{\operatorname{Imp}_b(P_j)\}$
\end{algorithmic}
\end{breakablealgorithm}
\begin{breakablealgorithm}
\caption{SMX: Graph construction and centrality computation}
\label{alg:smx_phase5}
\begin{algorithmic}[1]
\Require $\mathcal{P}$ (predicates), $\{\mathcal{L}_b\}_{b=1}^{B}$ (bag-wise ranked predicate lists), $\{\operatorname{Imp}_b(P_j)\}$ (normalized impacts), $\{\mathrm{VE}^{(m)}\}_{m=1}^{M}$ (explained variance ratios), seed $r$
\Ensure $G_r = (V,E)$ (directed weighted graph), $\{\mathrm{LRC}_r(P_j)\}$ (seed-specific predicate centralities)

\State Initialize directed weighted graph $G_r=(V,E)$
\State $V \leftarrow \mathcal{P} \cup \{\mathrm{Class}_0,\mathrm{Class}_1\}$
\For{$b = 1$ to $B$}
    \State Let $\mathcal{L}_b = [P_1^{(b)}, P_2^{(b)}, \dots, P_{L_b}^{(b)}]$
    \State Determine $c^* \leftarrow$ majority predicted class associated with the last ranked predicate in bag $b$
    \For{$l = 1$ to $L_b - 1$}
        \State Let $m_l$ be the zone associated with predicate $P_l^{(b)}$
        \State Set edge weight
        \[
            w_l \leftarrow \operatorname{Imp}_b(P_l^{(b)}) \times \mathrm{VE}^{(m_l)}
        \]
        \State Add or accumulate edge $(P_l^{(b)} \to P_{l+1}^{(b)})$ in $G_r$
    \EndFor
    \State Add terminal edge $(P_{L_b}^{(b)} \to \mathrm{Class}_{c^*})$ with weight
    \[
        w_{L_b} \leftarrow \operatorname{Imp}_b(P_{L_b}^{(b)}) \times \mathrm{VE}^{(m_{L_b})}
    \]
\EndFor
\For{each predicate node $P_j \in V$}
    \State Compute $\mathrm{LRC}_r(P_j)$ on $G_r$
\EndFor
\State \Return $G_r, \{\mathrm{LRC}_r(P_j)\}$
\end{algorithmic}
\end{breakablealgorithm}

On the other hand, the SMX computational cost depends on the number of samples $n$, the number of spectral variables $p$, the number of spectral zones $M$, the number of quantile levels $K=|\mathcal{Q}|$, the bag size $n_b$, the number of bags $B$, the number of retained predicates $N_P'$, and the $|\mathcal{R}|$ number of repetitions.

The spectral-zone decomposition and PCA aggregation stage requires fitting one one-component PCA model per zone, yielding complexity
\[
\mathcal{O}\!\left(\sum_{m=1}^{M} n d_m^2\right),
\]
where $d_m=|Z_m|$ is the size of zone $m$. For approximately balanced zones, this term becomes $\mathcal{O}(n p^2/M)$.

The predicate formulation stage computes thresholds over the zone scores, generates up to $2MK$ candidate predicates, removes duplicates, and constructs the indicator matrix. Its complexity is
\[
\mathcal{O}(nMK + nN_P'),
\]
which reduces to $\mathcal{O}(nMK)$ in the worst case, since $N_P' \leq 2MK$.

The dominant cost of SMX arises from stochastic bag generation and perturbation scoring. For each seed and each bag, the method evaluates the retained predicates, perturbs the associated spectral zones, and queries the explained model on original and perturbed samples. Let $C_f(n_b,p)$ denote the cost of evaluating the trained predictor on $n_b$ samples with $p$ variables, and let $d_{\max}$ be the maximum zone length. The complexity of this stage is
\[
\mathcal{O}\!\left(
|\mathcal{R}| \, B \, N_P'
\bigl[n_b d_{\max} + C_f(n_b,p)\bigr]
\right),
\]

The graph construction, centrality computation, and final ranking add lower-order terms dominated by sorting and graph traversal. Accordingly, the overall complexity of SMX can be summarized as
\[
\resizebox{\columnwidth}{!}{$
\mathcal{O}\!\left(
\sum_{m=1}^{M} n d_m^2
\;+\;
nMK
\;+\;
|\mathcal{R}| \, B \, N_P'
\bigl[n_b d_{\max} + C_f(n_b,p)\bigr]
\right)
$}
\]

In practice, the runtime is mainly governed by the perturbation stage, and therefore scales approximately linearly with the number of bags, retained predicates, and random seeds, while being strongly influenced by the inference cost of the explained model.


\subsection{Time cost comparison}
\label{app:timecost}

Table~\ref{tab:runtime_comparison} reports the wall-clock runtime (in seconds) of SMX, SHAP, and PFI for every model--dataset combination. Accordingly, the hierarchy of computational cost based on mean runtimes across datasets was SHAP $>$ SMX $>$ PFI for all models. Due to its reliance on \texttt{KernelExplainer}, SHAP is generally the most expensive of the three methods, especially given the high dimensionality of the spectral data, and the usage of all training samples as background data for the kernel estimation to ensure the most accurate approximation. Furthermore, although SMX and PFI were faster, their runtime can increase or decrease according to the adopted hyperparameters, \textit{i.e.}, the PFI number of repetitions and SMX ensemble strategy, as described in the Appendix~\ref{app:pseudocode}. VIP was excluded from the comparison because it is only defined for PLS and its computation is essentially instantaneous (it reads directly from the model's latent-variable decomposition). Overall, all methods remain practical for the dataset sizes considered, with runtimes ranging from tens of seconds to a few minutes.

\begin{table}[H]
  \centering
  \scriptsize
  \setlength{\tabcolsep}{3pt}
\caption{Runtime comparison (seconds) of SMX, PFI, and SHAP across models and datasets}
\label{tab:runtime_comparison}
 \resizebox{\linewidth}{!}{%
\begin{tabular}{lrrrrrrrrr}
  \toprule
   & \multicolumn{3}{c}{PLS} & \multicolumn{3}{c}{SVM} & \multicolumn{3}{c}{MLP} \\
   \cmidrule(lr){2-4} \cmidrule(lr){5-7} \cmidrule(lr){8-10}
  Dataset & SMX & PFI & SHAP & SMX & PFI & SHAP & SMX & PFI & SHAP \\
  \midrule
  Bank notes & 107.5 & 73.5 & 1692.8 & 104.1 & 98.6 & 10079.1 & 67.8 & 47.5 & 1540.1 \\
  Forage & 143.4 & 118.1 & 552.7 & 111.4 & 63.5 & 1600.1 & 115.1 & 58.4 & 487.9 \\
  Milk & 173.3 & 77.8 & 1513.6 & 142.0 & 244.0 & 28244.7 & 90.4 & 49.7 & 1363.5 \\
  Sediments & 177.3 & 53.2 & 54.9 & 86.5 & 53.6 & 103.1 & 90.7 & 52.8 & 49.7 \\
  Soil (GRS) & 50.5 & 14.6 & 42.6 & 29.0 & 16.3 & 151.9 & 29.4 & 13.7 & 43.0 \\
  Soil (XRF) & 82.8 & 37.7 & 289.9 & 72.6 & 39.6 & 2073.3 & 59.3 & 19.1 & 281.3 \\
  Soil types & 27.8 & 15.7 & 92.7 & 23.7 & 8.2 & 121.0 & 25.3 & 8.7 & 98.2 \\
  Synthetic & 54.7 & 43.1 & 1086.5 & 56.8 & 24.2 & 2292.2 & 36.9 & 16.0 & 517.3 \\
  Tomato & 142.4 & 44.1 & 52.7 & 71.4 & 46.6 & 121.2 & 73.9 & 43.9 & 47.7 \\
  \midrule
  \textbf{Mean} & \textbf{98.7} & \textbf{48.4} & \textbf{550.9} & \textbf{70.8} & \textbf{56.1} & \textbf{3943.4} & \textbf{60.4} & \textbf{31.5} & \textbf{424.3} \\
  \bottomrule
\end{tabular}
}
\end{table}


\subsection{Synthetic spectral data generation}
\label{subsec:synthetic_generation}
The synthetic data were generated by modeling each spectrum as a superposition of Gaussian peaks plus additive noise. Formally, each synthetic spectrum was described by
\begin{equation}\label{eq:synth_spectrum}
  S(x) = \sum_{i} A_i \exp\!\left(-\frac{(x - c_i)^2}{2\sigma_i^2}\right) + \varepsilon(x),
\end{equation}
where $c_i$, $A_i$ and $\sigma_i$ are, respectively, the center, amplitude and width (standard deviation) of the $i$-th Gaussian peak, and $\varepsilon(x)\sim\mathcal{N}(0,\,\sigma_\varepsilon^2)$ is a white-noise baseline. To introduce realistic sample-to-sample variability, both $A_i$ and $\sigma_i$ were drawn independently for each sample from normal distributions: $A_i \sim \mathcal{N}(\bar{A},\,s_A^2)$ and $\sigma_i \sim \mathcal{N}(\bar{\sigma},\,s_\sigma^2)$.

\subsection{Details on the real XRF and GRS datasets}
\label{subsec:real_datasets}

Table employed real datasets covered binary classification tasks across various application domains, including agriculture, food quality, and material authentication. Their details, as well as the objective of each classification task, are summarized in Table~\ref{tab:detailed_real_datasets}. 

\begin{table*}[]
  \centering
  \caption{Evaluated binary class XRF and GRS datasets and their descriptions}
  \label{tab:detailed_real_datasets}
  \resizebox{\textwidth}{!}{%
  \begin{tabular}{llcccccc}
    \toprule
    Dataset & Classification task & $n$ & Class A & Class B & $p$ & Range (keV) & $Zones$ \\
    \midrule
    Bank notes (XRF)   & Bank notes authentication           & 407 & 251 & 156 & 785  & 2.74-22.71 & 15 \\
    Forage (XRF)        & Forage provenance                   & 195 &  58 & 137 & 971  & 1.4-20.81  & 22 \\
    Milk (XRF)          & Whey protein adulterated milk       & 383 & 143 & 240 & 781  & 2.66-22.62 & 10 \\
    Soil fertility (XRF)& Soil fertility characterization     & 212 & 110 & 102 & 590  & 1.32-13.10 & 20 \\
    Soil fertility (GRS)& Soil fertility characterization     & 80 & 56 & 24 & 516  & 95-610 & 9 \\
    Soil types (GRS)    & Soil type                           & 156 &  77 &  79 & 374  & 1.32-13.10 & 21 \\
    Sediments (XRF)       & Sediments provenance              &  50 &  25 &  25 & 1166 & 1.40-13.05 & 19 \\
    Tomato (XRF)       & Tomato type                         &  52 &  20 &  32 & 1049 & 2.12-23.08 & 17 \\
    \bottomrule
  \end{tabular}
  \smallskip
  }
  \footnotesize{$n$: total number of samples; $p$: number of spectral variables; $Zones$: number of spectral zones}
\end{table*}

\subsection{Spectral zone definitions according to each dataset}
\label{subsec:zone_details}

The spectral zones used for each dataset are defined in the tables bellow. These zones were defined based on the known elemental composition of the samples, the expected spectral features such as characteristic peaks and scattering regions, and the experimental conditions (\textit{e.g.}, excitation energy, detector resolution). The zones named \textit{background} correspond to regions with no expected elemental signal and serve as uninformative reference intervals for the explainability analyses. The classifications followed well-documented XRF and GRS features (\citet{Beckhoff2006,VanGrieken2001,Knoll2010,Gilmore2008}), as well as dataset and instrument-specific characteristics.

\begin{table}[H]
\centering
\caption{Spectral zone definitions and plausibility for the Milk (\texttt{xrf}) 
dataset. Energy values are in keV}
\label{tab:zones_milk}
\begin{tabular}{lccc}
\toprule
\textbf{Zone} & \textbf{Start} & \textbf{End} & \textbf{Plausibility} \\
\midrule
Ag L$\alpha$ & 2.66 & 3.10 & Plausible \\
Ag L$\beta$ & 3.10 & 3.46 & Plausible \\
Ca & 3.46 & 3.92 & Plausible \\
background & 3.92 & 6.12 & Non-plausible \\
Fe & 6.12 & 6.68 & Plausible \\
Cu & 6.70 & 8.37 & Plausible \\
Zn & 8.37 & 9.10 & Plausible \\
Bremsstrahlung & 9.10 & 20.06 & Non-plausible \\
Ag Compton & 20.06 & 21.62 & Plausible \\
Ag k$\alpha$ & 21.48 & 22.62 & Plausible \\
\bottomrule
\end{tabular}
\end{table}

\begin{table}[H]
\centering
\caption{Spectral zone definitions and plausibility for the Bank 
Notes (\texttt{xrf}) dataset. Energy values are in keV}
\label{tab:zones_banknotes}
\begin{tabular}{lccc}
\toprule
\textbf{Zone} & \textbf{Start} & \textbf{End} & \textbf{Plausibility} \\
\midrule
Ar k$\alpha$ + Ag L & 2.76 & 3.47 & Plausible \\
Ca k$\alpha$ & 3.50 & 3.91 & Plausible \\
Ca k$\beta$ & 3.93 & 4.24 & Plausible \\
Ti k$\alpha$ & 4.26 & 4.72 & Plausible \\
Ti k$\beta$ & 4.75 & 5.13 & Plausible \\
background 1 & 5.16 & 6.12 & Non-plausible \\
Fe k$\alpha$ & 6.15 & 6.76 & Plausible \\
Fe k$\beta$ & 6.79 & 7.32 & Plausible \\
background 2 & 7.35 & 7.78 & Non-plausible \\
Cu k$\alpha$ & 7.81 & 8.29 & Plausible \\
Zn k$\alpha$ & 8.29 & 8.80 & Plausible \\
Cu k$\beta$ & 8.80 & 9.26 & Plausible \\
Zn k$\beta$ & 9.26 & 10.00 & Plausible \\
background 3 & 10.00 & 21.46 & Non-plausible \\
Ag k$\alpha$ scattering & 21.49 & 22.71 & Plausible \\
\bottomrule
\end{tabular}
\end{table}

\begin{table}[H]
\centering
\caption{Spectral zone definitions and plausibility for the 
Synthetic dataset. Values are in arbitrary units. The ground truth 
plausibility of each zone is determined by the data generation 
process described in Table~\ref{tab:detailed_real_datasets}: Feature~1, 
Feature~2, and Feature~3 correspond to spectral regions containing 
Gaussian peaks with class-discriminative amplitudes, while 
background zones contain only additive white noise with no 
systematic class-related signal}
\label{tab:zones_synthetic}
\begin{tabular}{lccc}
\toprule
\textbf{Zone} & \textbf{Start} & \textbf{End} & \textbf{Plausibility} \\
\midrule
Background 1 & 1.0 & 101.0 & Non-plausible \\
Feature 1 & 101.0 & 193.3 & Plausible \\
background 2 & 193.3 & 255.42 & Non-plausible \\
Feature 2 & 255.42 & 341.57 & Plausible \\
background 3 & 341.57 & 460.0 & Non-plausible \\
Feature 3 & 460.0 & 539.9 & Plausible \\
background 4 & 539.9 & 600.0 & Non-plausible \\
\bottomrule
\end{tabular}
\end{table}

\begin{table}[H]
\centering
\caption{Spectral zone definitions and plausibility for the Soil 
Types (\texttt{grs}) dataset. Channel numbers correspond to the acquired gamma-ray 
spectrum}
\label{tab:zones_soiltypes}
\begin{tabular}{lccc}
\toprule
\textbf{Zone} & \textbf{Start} & \textbf{End} & \textbf{Plausibility} \\
\midrule
Pb-212 & 57 & 70 & Plausible \\
Pb-214 & 70 & 115 & Plausible \\
Ac-228 + Tl-208 & 115 & 137 & Plausible \\
Tl-208 & 137 & 170 & Plausible \\
background 1 & 170 & 207 & Non-plausible \\
Ac-228 & 207 & 255 & Plausible \\
Bi-214 & 255 & 290 & Plausible \\
background 2 & 290 & 390 & Non-plausible \\
K-40 & 390 & 430 & Plausible \\
\bottomrule
\end{tabular}
\end{table}

\begin{table}[H]
\centering
\caption{Spectral zone definitions and plausibility for the Forage (\texttt{xrf})
dataset. Energy values are in keV}
\label{tab:zones_forage}
\begin{tabular}{lccc}
\toprule
\textbf{Zone} & \textbf{Start} & \textbf{End} & \textbf{Plausibility} \\
\midrule
Al & 1.40 & 1.63 & Plausible \\
Si & 1.63 & 1.86 & Plausible \\
P & 1.86 & 2.16 & Plausible \\
S & 2.16 & 2.44 & Plausible \\
Rh L + Ar & 2.44 & 3.10 & Plausible \\
K & 3.10 & 3.46 & Plausible \\
Ca k$\alpha$ & 3.46 & 3.86 & Plausible \\
Ca k$\beta$ & 3.86 & 4.37 & Plausible \\
Ti k$\alpha$ & 4.37 & 4.66 & Plausible \\
Ti k$\beta$ & 4.66 & 5.08 & Plausible \\
background 1 & 5.08 & 5.72 & Non-plausible \\
Mn & 5.72 & 6.10 & Plausible \\
Fe k$\alpha$ & 6.10 & 6.76 & Plausible \\
Fe k$\beta$ & 6.76 & 7.20 & Plausible \\
Ni & 7.20 & 7.69 & Plausible \\
Cu & 7.69 & 8.45 & Plausible \\
Zn & 8.45 & 8.81 & Plausible \\
background 2 & 8.81 & 13.10 & Non-plausible \\
Sum Fe & 13.10 & 13.63 & Non-plausible \\
background 3 & 13.63 & 18.00 & Non-plausible \\
Compton scattering & 18.00 & 19.70 & Plausible \\
Rayleigh scattering & 19.70 & 20.81 & Plausible \\
\bottomrule
\end{tabular}
\end{table}

\begin{table}[H]
\centering
\caption{Spectral zone definitions and plausibility for the Soil 
(\texttt{grs}) dataset. Channel numbers correspond to the acquired 
gamma-ray spectrum}
\label{tab:zones_soil_grs}
\begin{tabular}{lccc}
\toprule
\textbf{Zone} & \textbf{Start} & \textbf{End} & \textbf{Plausibility} \\
\midrule
Pb-212 & 95 & 115 & Plausible \\
Pb-214 & 115 & 190 & Plausible \\
Ac-228 + Tl-208 & 190 & 232 & Plausible \\
Tl-208 & 232 & 275 & Plausible \\
background 1 & 275 & 350 & Non-plausible \\
Ac-228 & 350 & 420 & Plausible \\
Bi-214 & 420 & 480 & Plausible \\
background 2 & 480 & 540 & Non-plausible \\
K-40 & 540 & 610 & Plausible \\
\bottomrule
\end{tabular}
\end{table}

\begin{table}[H]
\centering
\caption{Spectral zone definitions and plausibility for the Soil 
(\texttt{xrf}) dataset. Energy values are in keV}
\label{tab:zones_soil_xrf}
\begin{tabular}{lccc}
\toprule
\textbf{Zone} & \textbf{Start} & \textbf{End} & \textbf{Plausibility} \\
\midrule
Al & 1.33 & 1.63 & Plausible \\
Si & 1.63 & 1.86 & Plausible \\
P & 1.86 & 2.19 & Plausible \\
S & 2.19 & 2.55 & Plausible \\
Rh L + Ar & 2.55 & 3.21 & Plausible \\
K & 3.21 & 3.53 & Plausible \\
Ca k$\alpha$ & 3.53 & 3.84 & Plausible \\
Ca k$\beta$ & 3.84 & 4.37 & Plausible \\
Ti k$\alpha$ & 4.37 & 4.75 & Plausible \\
Ti k$\beta$ & 4.75 & 5.12 & Plausible \\
Cr & 5.12 & 5.77 & Plausible \\
Mn & 5.77 & 6.13 & Plausible \\
Fe k$\alpha$ & 6.13 & 6.80 & Plausible \\
Fe k$\beta$ & 6.80 & 7.30 & Plausible \\
background 1 & 7.30 & 7.91 & Non-plausible \\
Cu & 7.91 & 8.20 & Plausible \\
background 2 & 8.20 & 10.69 & Non-plausible \\
Fe k$\alpha$ + Ti k$\alpha$ & 10.69 & 11.14 & Non-plausible \\
background 3 & 11.14 & 12.55 & Non-plausible \\
Sum Fe & 12.55 & 13.10 & Non-plausible \\
\bottomrule
\end{tabular}
\end{table}

\begin{table}[H]
\centering
\caption{Spectral zone definitions and plausibility for the Tomato (\texttt{xrf}) 
dataset. Energy values are in keV}
\label{tab:zones_tomato}
\begin{tabular}{lccc}
\toprule
\textbf{Zone} & \textbf{Start} & \textbf{End} & \textbf{Plausibility} \\
\midrule
S & 2.12 & 2.40 & Plausible \\
Cl & 2.40 & 3.04 & Plausible \\
K k$\alpha$ & 3.00 & 3.44 & Plausible \\
K k$\beta$ + Ca k$\alpha$ & 3.44 & 3.84 & Plausible \\
background 1 & 3.84 & 5.70 & Non-plausible \\
Mn & 5.70 & 6.04 & Plausible \\
Fe & 6.04 & 7.30 & Plausible \\
Cu & 7.30 & 8.32 & Plausible \\
Zn & 8.32 & 8.84 & Plausible \\
background 2 & 8.84 & 11.56 & Non-plausible \\
Br k$\alpha$ & 11.56 & 12.50 & Plausible \\
Rb k$\alpha$ + Br k$\beta$ & 12.50 & 13.80 & Plausible \\
background 3 & 13.80 & 17.00 & Non-plausible \\
Rh k$\alpha$ Compton & 17.00 & 19.90 & Plausible \\
Rh k$\alpha$ Rayleigh & 19.90 & 20.80 & Plausible \\
Rh k$\beta$ Compton & 20.80 & 22.30 & Plausible \\
Rh k$\beta$ Rayleigh & 22.30 & 23.08 & Plausible \\
\bottomrule
\end{tabular}
\end{table}

\begin{table}[H]
\centering
\caption{Spectral zone definitions and plausibility for the 
Sediments dataset. Energy values are in keV}
\label{tab:zones_sediments}
\begin{tabular}{lccc}
\toprule
\textbf{Zone} & \textbf{Start} & \textbf{End} & \textbf{Plausibility} \\
\midrule
Al & 1.40 & 1.61 & Plausible \\
Si & 1.61 & 1.93 & Plausible \\
P & 1.86 & 2.16 & Plausible \\
S & 2.16 & 2.52 & Plausible \\
Rh L + Ar & 2.52 & 3.15 & Plausible \\
K & 3.15 & 3.51 & Plausible \\
Ca k$\alpha$ & 3.51 & 3.88 & Plausible \\
Ca k$\beta$ & 3.88 & 4.18 & Plausible \\
Ti k$\alpha$ & 4.18 & 4.77 & Plausible \\
Ti k$\beta$ & 4.77 & 5.15 & Plausible \\
background 1 & 5.15 & 5.74 & Non-plausible \\
Mn & 5.74 & 6.11 & Plausible \\
Fe k$\alpha$ & 6.11 & 6.76 & Plausible \\
Fe k$\beta$ & 6.76 & 7.36 & Plausible \\
background 2 & 7.36 & 7.95 & Plausible \\
Cu & 7.95 & 8.25 & Plausible \\
Zn & 8.25 & 8.92 & Plausible \\
background 3 & 8.92 & 12.71 & Non-plausible \\
sum Fe & 12.71 & 13.05 & Non-plausible \\
\bottomrule
\end{tabular}
\end{table}

\newpage
\subsection{Structural Comparison Between SMX and DPG}
\label{appem:smx_dpg}

This section provides a detailed structural comparison between the Spectral Model eXplainer (\texttt{smx}) proposed in this work and the Decision Predicate Graph (\texttt{dpg}; \citep{arrighi2024dpg}), which served as the foundational architecture for \texttt{smx}. 

\begin{table}[H]
\centering
\caption{Structural comparison between the Decision Predicate Graph (\texttt{dpg}; \citep{arrighi2024dpg}) and the Spectral Model eXplainer (\texttt{smx}) proposed in this work. \texttt{smx} inherits the predicate-graph architecture of \texttt{dpg} and extends it with spectral zone aggregation, \texttt{pca}-based threshold reconstruction, variance-explained edge weighting, and zone-length normalization, all motivated by properties specific to spectroscopic data.}
\label{tab:smx_vs_dpg}
\renewcommand{\arraystretch}{1.4}
\begin{tabularx}{\linewidth}{@{} >{\raggedright\arraybackslash}p{2.2cm} X >{\raggedright\arraybackslash}p{3.2cm} @{}}
\toprule
\textbf{Property} 
    & \textbf{DPG \citep{arrighi2024dpg}} 
    & \textbf{SMX} \\
\midrule
Target models
    & Tree ensembles
    & Any trained classifier (model-agnostic) \\

Explanatory unit
    & Individual input features
    & Expert-defined spectral zones \\

Pre-graph aggregation
    & None (raw features)
    & \texttt{pca}-based zone score compression \\

Predicate basis
    & Decision node splits
    & Quantile thresholds on zone \texttt{pca} scores \\

Edge weighting
    & Co-occurrence of samples satisfying predicate pairs
    & Perturbation impact $\times$ explained variance ratio \\

Impact normalisation
    & None
    & Normalised by zone length \\

Output
    & Ranked predicate list $+$ graph
    & Ranked predicate list $+$ graph $+$ threshold spectrum \\

Threshold back-projection
    & Not available
    & Full spectral profile in natural units (Eq.~\ref{eq:threshold_spectrum}) \\

Stability mechanism
    & Single run
    & Multi-seed averaging over $R$ repetitions \\

Domain
    & General tabular data
    & Spectroscopic measurements \\
\bottomrule
\end{tabularx}
\end{table}

\end{document}